\documentclass[dvips,aos,preprint]{imsart}

\RequirePackage[OT1]{fontenc}
\RequirePackage{amsthm,amsmath}
\usepackage{natbib}
  \bibpunct{(}{)}{;}{a}{}{,} 
\RequirePackage[colorlinks,citecolor=blue,urlcolor=blue]{hyperref}

\usepackage{amsmath}
\usepackage{amsbsy}
\usepackage{amscd}
\usepackage{amsopn}
\usepackage{amstext}
\usepackage{amsxtra}
\usepackage{epsfig}
\usepackage{times} 

\usepackage{epsfig}

\usepackage{fancyhdr}
\usepackage{graphicx}
\newtheorem{theorem}{Theorem}

\newtheorem{corollary}{Corollary}
\newtheorem{lemma}{Lemma}

\newtheorem{prop}{Proposition}

\usepackage{amsfonts}
\usepackage{amssymb}
\usepackage{latexsym}
\usepackage{mathrsfs}
\usepackage[all]{xy}
\usepackage{graphicx}
\usepackage{amsmath}
\usepackage[all]{xy}

\def\lesssim{\mathrel{\hbox{\rlap{\hbox{\lower4pt\hbox{$\sim$}}}\hbox{$<$}}}}

\def\corref#1{}

\startlocaldefs
\numberwithin{equation}{section}
\theoremstyle{plain}

\endlocaldefs

\begin{document}

\begin{frontmatter}
\title{Inference of network summary statistics\\ through nonparametric network denoising\thanksref{T1}}
\runtitle{Inference of Network Summary Statistics Through Denoising}
\thankstext{T1}{This work was supported, in part, by AFOSR award 12RSL042 and NSF grant CNS-0905565 to Boston University, and by ARO MURI award W911NF-11-1-0036 and NSF CAREER grant IIS-1149662 to Harvard University. Edoardo M. Airoldi is an Alfred P. Sloan Research Fellow. This work begun during the Program on Complex Networks hosted by SAMSI in 2010--2011.}
\thankstext{c1}{Corresponding Author.}

\begin{aug}
\author{\fnms{Prakash} \snm{Balachandran}\thanksref{m1}\ead[label=e1]{prakashb@bu.edu}\ead[label=u1,url]{http://math.bu.edu/people/prakashb/}},
\author{\fnms{Edoardo M.} \snm{Airoldi}\thanksref{m2}
\ead[label=e3]{airoldi@fas.harvard.edu}
\ead[label=u3,url]{http://www.fas.harvard.edu/$\sim$airoldi/}}\\
\and
\author{\fnms{Eric D.} \snm{Kolaczyk} \corref{c1} \thanksref{m1,c1}\ead[label=e2]{kolaczyk@math.bu.edu}\ead[label=u2,url]{http://math.bu.edu/people/kolaczyk/}}\\

\runauthor{P. Balachandran et al.}

\affiliation{$^\ddag$Boston University and $^\S$Harvard University}

\address{\thanksmark{m1}
Boston University\\
Department of Mathematics and Statistics\\
111 Cummington Street, Boston, MA 02215, USA\\
\printead{e1}\\
\printead{u1}\\
\printead{e2}\\
\printead{u2}\\}

\address{\thanksmark{m2}
Harvard University\\
Department of Statistics\\
1 Oxford Street\\
Cambridge, MA 02138, USA\\
\printead{e3}\\
\printead{u3}}
\end{aug}

\begin{abstract}
Consider observing an undirected network that is `noisy' in the sense that there are Type I and Type II errors in the observation of edges. Such errors can arise, for example, in the context of inferring gene regulatory networks in genomics or functional connectivity networks in neuroscience. Given a single observed network then, to what extent are summary statistics for that network representative of their analogues for the true underlying network? Can we infer such statistics more accurately by taking into account the noise in the observed network edges? 
In this paper, we answer both of these questions.  In particular, we develop a spectral-based methodology using the adjacency matrix to `denoise' the observed network data and produce more accurate inference of the summary statistics of the true network.  We characterize  performance of our methodology through bounds on appropriate notions of risk in the $L^2$ sense, and conclude by illustrating the practical impact of this work on synthetic and real-world data.
\end{abstract}



\end{frontmatter}

\tableofcontents 

\section{Introduction}
\label{1}

Driven by applications in the natural, physical and information sciences, statistical models, methods and theory for analyzing network data have received increasing attention in the past decade \citep{Alon:2006fk,Jack:2008,Newman:2010uq,Goldenberg_Zheng_Fienberg_2009,kola:2009}. 
Arguably, one of the most common paradigms in network analysis consists of the following steps:
(i) collecting measurements on a complex system of interest, 
(ii) constructing a network representation using these measurements, and 
(iii) characterizing the structure of the network through various network summary statistics.  
Importantly, errors in the original measurements induce structured errors in the network, and hence uncertainty associated with summaries of the network.  
In this paper, we explore  the effects of error propagation from raw measurements to network representation, to summary statistics. We develop a nonparametric strategy to control inferential errors for a number of summary statistics, we characterize the way in which error levels and the underlying network structure impact the performance of the proposed strategy, and we illustrate how our theory can inform applications to gene regulatory networks and protein-protein interaction networks.  

In a number of applications, the step of network construction is frequently approached as an inferential task, in which with the goal is making inference on the edges (and possibly the edge weights) in a true, underlying network.\footnote{See~\citet[Ch 7.3]{kola:2009} for an overview of the statistical approaches to this problem.} Importantly, in the process of inferring such a network, errors of both Type I, i.e., declaring an edge where none exists, and Type II, i.e., omitting an edge when it exists, can be expected to happen.  
Applications where network data are expected to contain errors due to errors in the raw measurements include the analysis of gene regulatory networks using microarray data (e.g., \cite{emmert2012statistical,lee2009computational,Pham:2011ys,Finegold:2011zr}), 
 protein-protein interaction networks using mass spectrometry or tandem affinity purification data (e.g., \cite{Airo:Blei:Fien:Xing:2006c,PPI,Telesca:2012vn}), 
and various related ``-omics'' networks constructed from multiple simultaneous measurements
in systems biology (e.g., \cite{Katenka:2012kx}), as well as 
 functional connectivity networks using fMRI data  (e.g., \cite{smith2011network,Priebe:2012ly}).  

After constructing the network, scientists are typically most interested in characterizing it, quantitatively.  Numerical summaries of all sorts are used to this end, which range from vertex degrees to more general notions of vertex `importance', often called centrality measures, to more global measures of network coherence, such as the density of edges or the extent to which `flows' can move throughout the network, often summarized through the so-called conductance.  See, for example, \cite[Ch 4]{kola:2009}, for an overview.

However, such network summaries are almost always presented without any quantification, nor acknowledgement, of  the uncertainty  they inherit from the inferential procedure in the network construction step.  
The most likely reason for this omission would appear to be the lack of tools and techniques for understanding the propagation of error, from measurements to network summary statistics.  
To the best of our knowledge, the inferential issues we have raised have not even received a formal treatment to date.

The primary contribution of this paper is to study a general formulation of this problem and, within this context, to characterize the performance of certain estimators of network summaries in terms of bounds on their statistical risk.
Here we are concerned with undirected, $\{0,1\}$-valued, networks represented by means of the corresponding adjacency matrix, $W$. We assume that we are given a single noisy instance 
\[
 W_{obs}=W_{true}+W_{noise}
\] 
of some true network $W_{true}$.  For a given a statistic $g(W)$ on the network, our goal is to construct an estimator $\widehat{g}$ of $g(W_{true})$ such that 
\[
 \mathbb{E}\left[\, (\widehat{g}-g\left(W_{true}\right))^2\, \right] \ll \mathbb{E}\left[\, (g\left(W_{obs}\right)-g\left(W_{true}\right))^2\, \right].
\] 

The approach we develop to find $\widehat{g}$ is rooted in the literature on a related problem in signal processing -- that of nonparametric function estimation, or `denoising'.  
There, given a single instance of 
\[
 f_{obs}=f_{true}+f_{noise},
\] 
the goal is to construct an estimator $\widehat{f}$ so that 
\[
 \mathbb{E}||\widehat{f}-f_{true}||^2_2 \ll \mathbb{E}||f_{obs}-f_{true}||^2_2 \enskip ,
\]
where $||\cdot||_2$ refers to the $L^2$ norm measuring the mean-square error.  
A classical approach to this problem is to assume that the true signal $f_{true}$ is compressible in some set of basis functions, e.g., a Fourier or wavelet orthonormal basis \citep[e.g., see][]{Strang:1996fk}.  The estimate $\widehat{f}$ is then constructed by projecting $f_{obs}$ onto this basis and thresholding the corresponding coefficients, discarding higher `frequency' components, and typically obtaining a much better reconstruction of $f_{true}$ than $f_{obs}$, under suitable assumptions on $f_{true}$.

We adopt a similar strategy here, based on spectral analysis of the network, through its adjacency matrix $W_{obs}$ \citep{Chung:1996kx,Bai:2010uq}.  Specifically, in our approach, we construct an estimator $\widehat{W}$ of $W_{true}$ by projecting the noisy observation $W_{obs}$ onto a carefully chosen basis containing most of the energy of the original signal, $W_{true}$.  The compression of information of $W_{true}$ in this basis allows one to keep some of the projections while discarding the remaining ones, resulting in an improved reconstruction of the original signal with minimal error.  We find that for sufficiently smooth summaries $g$, the accuracy of estimation of $W_{true}$ by $\widehat{W}$ translates to accuracy in estimation of $g(W_{true})$ by $g(\widehat{W})$. 

The organization of this paper is as follows.  In Section~\ref{2}, we provide background and notation, discussing assumptions on the noise $W_{noise}$ and sketching our overall approach.  In Section~\ref{three}, we present our main results: (i) characterization of $W_{obs}$ and $g(W_{obs})$ as naive estimators, (ii) characterization of our proposed $\widehat{W}$ and $g(\widehat{W})$, in comparison, and (iii) a partial characterization of common choices of network summary statistics to which our method does and does not pertain.  Finally, in Section \ref{4}, we evaluate our methods on simulated data and we  demonstrate how our theory can inform data analysis in applications to protein interaction networks and gene regulatory networks.

\section{Background and approach}
\label{2}

\subsection{Background and notation}
\label{2.1}

Let $G_{true}=(V_{true},E_{true})$ be a (fixed) graph representing an underlying network of interest, defined on $n=|V_{true}|$ vertices with $m=|E_{true}|$ edges.  We assume that $G_{true}$ is undirected, and has no multi-edges or self-loops.  This last condition is not necessary, but makes certain computations such as the moments of the random spectral radius, easier.  See Section \ref{three} for details.  Denote by $W_{true}$ the $n\times n$ adjacency matrix for $G_{true}$, where $W_{ij}=1$ if $(i,j)\in E_{true}$, and zero otherwise.  

Similarly, let $G_{obs}$ represent a version of $G_{true}$ observed with noise, and $W_{obs}$, the corresponding $n\times n$ adjacency matrix.  We specify the precise nature of this noise through the form of the adjacency matrix, writing
\begin{equation}
\label{eq:the.model}
W_{obs}=W_{true}+W_{noise} \enskip ,  
\end{equation}
where $W_{noise}$ is an additive noise such that
\begin{enumerate}

\item  $W_{noise}(i,j)\sim -Bern(p)$, if $W_{true}(i,j)=1$;

\item  $W_{noise}(i,j)\sim Bern(q)$, if $W_{true}(i,j)=0$.
\end{enumerate}
Here $q$ can be interpreted as the probability of a Type I error (false positive), and $p$, as the probability of a Type II error (false negative), which are the same fixed values across all edges.  These assumptions of homogeneous Type I and II error probabilities (i.e. across edges) allow for the construction of an unbiased version of $W_{obs}$, denoted $\widetilde{W}_{obs}$ (see Theorem \ref{Raw Data}), which in turn will prove useful in producing interpretable bounds on the accuracy of estimation of $W_{true}$ by $\widehat{W}$.  These assumptions are not necessary, but rather are a simplifying feature of our analysis.

We note that we have made no assumptions on the dependency of the noise.  Indeed, our work can accommodate dependency under suitable assumptions on its effect on the observed maximum degree squared and on the correlation coefficients of edge observations (see Section \ref{3.3} for details).  To date, however, the precise nature of this and similar dependencies arising in practice is largely unknown, and we therefore leave the exploration of that topic for future work.

We will represent a (real-valued) summary statistic of a network as a function $g$ applied to an adjacency matrix $W$.  Since our approach is one of spectral-based estimation of $W_{true}$, we are interested in continuous, particularly (locally) Lipschitz continuous, statistics since it allows us to control the accuracy in estimating $g\left(W_{true}\right)$ by the accuracy in estimating $W_{true}$.  More formally, we will restrict our attention to the class of Lipschitz statistics $g(W)$, for which
\begin{equation}
|g(W_1)-g(W_2)|\leq C||W_1-W_2||_1 \enskip ,
\label{eq:Lip}
\end{equation}
where $||\cdot||_1$ denotes the $L_1$ norm, i.e.,
$$||A||_1=\sum_{i,j}|A(i,j)|\enskip$$ and the constant $C$ can depend on $n$.
Hence, for sufficiently close $W_1$ and $W_2$, the values $g(W_1)$
and $g(W_2)$ will be close as well.  

As a result of this last fact, we will approach the study of estimators $\widehat g$ of $g$ through the study of estimators $\widehat{W}$ of $W_{true}$, with $$\widehat g =g(\widehat{W}).$$  We will evaluate the loss in estimating $g(W_{true})$ by $g(\widehat W)$ using squared-error loss, and analogously, we will evaluate the loss in estimating $W_{true}$ by $\widehat{W}$ using the Frobenius norm, i.e.,
$$||A||_F^2=\sum_{i,j}|A(i,j)|^2 .$$
As shall be seen, our rationale for choosing this norm is that it equals the sum square of eigenvalues for symmetric matrices, which upon truncation at a certain eigenvalue we will use as a complexity parameter in a bias-variance tradeoff.

We normalize the corresponding risks by appropriate powers of the order $n$ of the network graph $G_{true}$.  Because \citep{r1}
\begin{equation}
\label{LipL1L2}
\left| g(W_1)-g(W_2)\right|\leq C||W_1-W_2||_1\leq C\cdot n \cdot ||W_1-W_2||_F \enskip ,
\end{equation}
we have that
\begin{equation}
\label{Mean-squared}
\frac{\left\{\mathbb{E} \left[ g\left(\widehat{W}\right) - 
		g\left(W_{true}\right) \right]^2 \right\}^{\frac{1}{2}}}{n^2}
  \le  C\,  \frac{ \left\{\mathbb{E}\left[ ||\widehat{W} - W_{true} ||_F^2\right] \right\}^{\frac{1}{2}} }{n}\enskip ,
  \end{equation}
which suggests normalization by a factor of $n^{-2}$ on the scale of $g$, and of $n^{-1}$, on the scale of $W$.  For notational convenience, we write for two (random) $n\times n$ matrices $A$ and $B$, $d(A,B)=\left(\mathbb{E}\left[||A-B||_F^2\right]\right)^{1/2}$.

It might seem strange that while the natural setting for Lipchitz continuity of statistics $g$ is given with respect to the $L_1$ norm on $W$ in (\ref{eq:Lip}), we measure the mean-squared error of $g$ with respect to the mean-squared error of $W$.  Indeed, comparing (\ref{LipL1L2}) and  (\ref{Mean-squared}), one could argue that our scaling could be improved by keeping the $L_1$ norm without passing to the Frobenius norm.  However, since our approach is spectral with symmetric matrices, the Frobenius norm is invariant under orthogonal transformations while the $L_1$ norm is not.  Under an orthogonal base change then, errors of estimators using the Frobenius norm are unchanged, while they undergo distortions with respect to the $L_1$ norm.

\subsection{Spectral-based approach using $W$}
\label{2.2}

We now outline our approach of construction of $\widehat{W}$ mentioned in the previous section.

First, recall that the adjacency matrix $W=[W(i,j)]$ can be thought of as a linear operator on $\mathbb{R}^n$ represented in the natural basis $\Phi=\{1_v\}_{v\in V}$, consisting of indicator functions on vertices.  To uniquely define this operator, however, we can choose any basis.  

If $P$ is an $n\times n$ matrix whose columns form a new orthonormal basis for $\mathbb{R}^n$, then in the notation of Section \ref{2.1}, $$\widetilde{W}_{obs,P}=P^T\widetilde{W}_{obs}P, \; \; \; \widetilde{W}_{obs,P}(i,j)=\left<P(\cdot,i),\widetilde{W}_{obs}(\cdot,j)\right>$$  is the matrix representation of $\widetilde{W}_{obs}$ in the basis $\mathcal{B}=\{P(:,v)\}_{v\in V_{true}}$ and similarly for $W_{true,P}$.

So, in the norm $d(A,B)=\left(\mathbb{E}\left[||A-B||^2_F\right]\right)^{1/2}$, if we choose to keep entries $(i,j)\in S$ of $\widetilde{W}_{obs,P}$ and transform back, obtaining the estimator $\widehat{W}_{P,S}$, then

\begin{equation}
\begin{aligned}
\label{base change}
d\left(\widehat{W}_{P,S},W_{true}\right)^2&=\sum_{(i,j)\in S} \mathbb{E} \left\{ \left(\widetilde{W}_{obs,P}(i,j)-W_{true,P}(i,j)\right)^2\right\}\\
&\hspace{1.5in}+\sum_{(i,j)\in S^c} \mathbb{E} \left[W_{true,P}(i,j) (i,j)^2 \right]\\
&=\sum_{(i,j)\in S} \mathbb{E} \left\{ \left(\widetilde{W}_{obs,P}(i,j)-W_{true,P}(i,j)\right)^2\right\}\\
&\hspace{1.5in} + \sum_{(i,j)\in S^c} \left<P(\cdot, i),W_{true} P(\cdot, j)\right>^2\\
\end{aligned}
\end{equation}

Note three things about this representation:
\begin{enumerate}
\item  It holds in the case of independent and dependent noise.

\item  Taking $S$ to be all pairs of indices $(i,j)$, $$d\left(\widehat{W}_{P,S},W_{true}\right)^2=d(\widetilde{W}_{obs,P},W_{true,P})^2=d\left(\widetilde{W}_{obs},W_{true}\right)^2\enskip ,$$ since the Frobenius norm is invariant under orthogonal transformations.  Thus, we achieve no improvement over $\widetilde{W}_{obs}$.  

\item  The form of (\ref{base change}) reflects a trade off between the variance and bias terms, the first and second terms in (\ref{base change}), respectively.
\end{enumerate}

 In choosing a basis then, we want to simultaneously minimize both terms in (\ref{base change}).  Thus, we want:
\begin{enumerate}
\item  $\mathbb{E} \left\{ \left(\widetilde{W}_{obs,P}(i,j)-W_{true,P}(i,j)\right)^2\right\}$ to be large for $(i,j)\in S^c$, so that the projection of the noise onto these subspaces is large, so we can discard them;

\item  $\left<P(\cdot, i),W_{true} P(\cdot, j)\right>^2$ for $(i,j)\in S^c$ to be as small as possible, so that the bulk of the signal lives on the subspace corresponding to projections $(i,j)\in S$;

\item  For practical considerations, we want to minimize the number of such projections to compute, i.e., $|S|$.
\end{enumerate}

Ideally, a natural way to achieve these goals is by choosing the columns of $P$ to be the first few dominant eigenvectors of $W_{true}$, assuming that its squared eigenvalues are ordered descending.  That way, both (2) and (3) in the above list are satisfied, and we need only verify (1).  In addition, we show that, while this basis will not be known in practice, one can instead use the first few dominant eigenvectors of $\widetilde{W}_{obs}$ to approximate the corresponding dominant eigenvectors of $W_{true}$, at the cost of incurring an additional error term that under suitable conditions on the noise and the size of the network tends to zero as $n\rightarrow \infty$.  In particular, we will be interested in the limit of large $n$ with $p,q\sim O(1)$, but our results allow for more general considerations on $p$ and $q$.  

In the context of statistical estimation, we have in comparison to (\ref{Mean-squared}) that

\begin{equation}
\label{statistical estimation}
\begin{aligned}
\frac{\left\{\mathbb{E} \left[ g\left(\widehat{W}_{P,S}\right) - 
		g\left(W_{true}\right) \right]^2 \right\}^{\frac{1}{2}}}{n^2}
  &\leq  C  \frac{d\left(\widehat{W}_{P,S},W_{true}\right)}{n}\\
  & =C \frac{d\left(\widehat{W}_{P,S},W_{true}\right)}{d\left(\widetilde{W}_{obs},W_{true}\right)}\cdot \frac{d\left(\widetilde{W}_{obs},W_{true}\right)}{n}
\enskip .
  \end{aligned}
\end{equation}
Thus, the quantity $d\left(\widehat{W}_{P,S},W_{true}\right) / d\left(\widetilde{W}_{obs},W_{true}\right)$ defines the error of our estimator relative to the raw data, and when less than 1, determines how much better our estimator is guaranteed to perform in estimating statistics $g(\cdot)$ over the raw data $W_{obs}$.  As we will see in Theorem \ref{Raw Data}, $d\left(\widetilde{W}_{obs},W_{true}\right)/ n \sim O(1)$, and so the focus of the paper will be on bounding the relative error.

\section{Main results}
\label{three}

In this section, we present our main results.  
First, we define our estimator $\widehat{W}$ of $W_{true}$.  We then outline our program in bounding the relative error in estimating $W_{true}$, explaining its rationale. Finally, we proceed to a systematic analysis.

As a preliminary step in defining our estimator $\widehat{W}$, we define
$$\widetilde{W}_{obs}=\frac{W_{obs}-qW_{K_n}}{1-(p+q)}\enskip ,$$ 
where $W_{K_n}$ is a matrix of ones with zero diagonals.  The matrix $\widetilde{W}_{obs}$ is a centered and rescaled version of the original $W_{obs}$, and will be seen (i.e., Theorem~\ref{Raw Data} below) to be an unbiased estimator of $W_{true}$.  This step may be viewed as a pre-processing step in the overall definition of our estimator and assumes, in principle, knowledge of the Type I and II error probabilities $q$ and $p$. This assumption is made largely for theoretical convenience and improved interpretability of results, and is roughly analogous to assuming that one knows the noise level in a standard regression model, a not-uncommon assumption in theoretical calculations for regression.  

In practice, of course, the values of $q$ and $p$ must be obtained from context.  In Section~\ref{4} we present examples of how this may be done in two real-world settings, through (i) the use of experimentally reported error probabilities in the context of protein-protein interaction networks, and (ii) the estimation of error probabilities, using empirical null principles, in the context of gene regulatory networks inferred from microarray expression profiles.

Now let  $\{\phi_i, \mu_i\}_{i=1}^n$ be the eigensystem of $\widetilde{W}_{obs}$, ordered descending in $\{\mu_i^2\}_{i=1}^n$.  Define our estimator of $W_{true}$, using the first $s$ modes of $\widetilde{W}_{obs}$, to be $$\widehat{W}_s=\sum_{i=1}^s \left<\phi_i,\widetilde{W}_{obs}\phi_i\right>\phi_i\phi_i^T.$$

Our program in bounding the relative error of $\widehat{W}_s$ in estimating $W_{true}$ is as follows.

\begin{enumerate}

\item  {\em $s^{th}$ order spectral projection: ideal basis}

Let $\{\psi_i,\lambda_i\}_{i=1}^n$ be the eigensystem of $W_{true}$ ordered decreasing in $\{\lambda^2_i\}_{i=1}^n$, and define $$\widehat{W}_{ideal,s}=\sum_{i=1}^s \left<\psi_i,\widetilde{W}_{obs}\psi_i\right> \psi_i\psi_i^T.$$  We will bound $$d\left(\widehat{W}_{ideal,s},W_{true}\right)$$ where $d$ is the norm on $n\times n$ matrices, $d(A,B)=\sqrt{\mathbb{E}\left[||A-B||_F^2\right]}.$\\

\item  {\em $s^{th}$ order spectral projection: empirical basis}

Then, we will proceed to bound  $$d\left(\widehat{W}_s,\widehat{W}_{ideal,s}\right)$$ so that the final bound on the relative error of $\widehat{W}_s$ then becomes

\begin{equation}
\label{triangle estimate}
\frac{d\left(\widehat{W}_s,W_{true}\right)}{d\left(\widetilde{W}_{obs},W_{true}\right)} \leq \frac{d\left(\widehat{W}_s,\widehat{W}_{ideal,s}\right)}{d\left(\widetilde{W}_{obs},W_{true}\right)} + \frac{d\left(\widehat{W}_{ideal,s},W_{true}\right)}{d\left(\widetilde{W}_{obs},W_{true}\right)}.
\end{equation}

\end{enumerate}

The rationale for this program is as follows.  We call $\widehat{W}_{ideal,s}$ the $s^{th}$ order ideal estimator of $W_{true}$ since it assumes that we have access to the true eigenbasis of $W_{true}$.  As mentioned in Section \ref{2}, there is a bias-variance tradeoff due to the choice of the parameter $s$, and up to this choice, we show in Theorem \ref{Ideal Estimator} that the relative error of this ideal estimator is small.

In practice, however, we do not have the eigenbasis of $W_{true}$.  But since the relative error in estimating $W_{true}$ by $\widetilde{W}_{ideal,s}$ is small up to the choice of $s$, the remaining part of the relative error of $\widehat{W}_s$ to $W_{true}$ is determined by estimating $\widehat{W}_{ideal,s}$ by $\widehat{W}_s$.  Both of these estimators use the same $\widetilde{W}_{obs}$, but project them in directions determined by the eigenbases of $\widetilde{W}_{obs}$ and $W_{true}$.  This part of the relative error then is completely determined by the alignment of the eigenspaces of $\widetilde{W}_{obs}$ and $W_{true}$, and Theorem \ref{Empirical Estimator} implies that under suitable assumptions on $p,q,n$ and the maximum degree of $W_{true}$, this error is small.

The next few sections state our results in the above program.  First, in Section~\ref{3.1}, we characterize the performance of the naive estimator $g\left(\widetilde{W}_{obs}\right)$ in estimating $g\left(W_{true}\right)$.  Specifically, we provide exact expressions for how well we can do in mean-square error using just the raw data $W_{obs}$.  Second, in Section~\ref{3.2}, we bound $d\left(\widehat{W}_{ideal,s},W_{true}\right)$ and show that it is possible to do substantially better than the naive estimator, depending on the structure of the underlying network graph $G_{true}$, and discuss situations where this could occur.  Third, in Section~\ref{3.3}, we bound $d\left(\widehat{W}_{s},W_{true}\right)$.  Finally, in Section~\ref{3.4}, we present a list of various network summary statistics, characterized as to whether their corresponding functions $g$ are or are not Lipshitz, thus indicating to which choices of summary in practice our results pertain.

\subsection{Performance of the na\"ive estimator}
\label{3.1}

In practice, network summary statistics $g(W_{true})$ are frequently estimated through a simple plug-in estimator, i.e., using $g(W_{obs})$.  The following theorem characterizes the mean-square error performance of this estimator, appropriately renormalized.

\begin{theorem}
\label{Raw Data}
Under the model $W_{obs}=W_{true}+W_{noise}$ defined in (\ref{eq:the.model}), define 
$$p=\mathbb{P}\left[W_{noise}(i,j)=-1 | W_{true}(i,j)=1\right]$$ 
and 
$$q=\mathbb{P}\left[W_{noise}(i,j)=1 | W_{true}(i,j)=0\right]$$  
as the Type I and Type II errors, respectively, for all $1\leq i <j\leq n$.  Furthermore, let $$\widetilde{W}_{obs}=\frac{W_{obs}-qW_{K_n}}{1-(p+q)}$$ where $K_n$ is the all-ones matrix with zero diagonals.  Then,
\begin{enumerate}
\item  $\mathbb{E}\left[\widetilde{W}_{obs}\right]=\mathbb{E}\left[W_{true}\right]$ in any fixed basis.

\item  The expected mean square error of $\widetilde{W}_{obs}$ is,
\begin{equation}
\label{eq:mse.naive}
d\left(\widetilde{W}_{obs},W_{true}\right)^2 =\frac{2\left(p(1-p)m + q(1-q) \left[ {n\choose 2}-m\right]\right)}{(1-(p+q))^2}\\\enskip ,
\end{equation}
so that for any Lipschitz continuous statistic $g$ (see equation \ref{Mean-squared})
\begin{equation}
\label{eq:mse.g.naive}
\begin{aligned}
&n^{-4}\mathbb{E}\left\{[g(\widetilde{W}_{obs})-g(W_{true})]^2\right\}\\
&\leq 2\frac{C^2}{n^2} \frac{\left(p(1-p)m+q(1-q)\left[{n\choose 2}-m\right] \right)}{(1-(p+q))^2}\enskip .
\end{aligned}
\end{equation}
\end{enumerate}
\end{theorem}

The proof of this theorem is given in Appendix \ref{A}.  Note that since ${n\choose 2}\sim n^2/2$, the right-hand side of (\ref{eq:mse.g.naive}) behaves like 
$$p(1-p) \cdot {\rm den}(G_{true}) + q(1-q) \cdot  \left( 1- {\rm den}(G_{true}) \right) \enskip ,$$
where $\hbox{den}(G_{true}) = m / {n\choose 2}$ is the density of the graph $G_{obs}$.  Hence, the performance of the naive estimator is driven by a combination of (i) the Type I and II error rates, $q$ and $p$ respectively, and ii) the network structure, i.e. via the edge density.

It has been observed in practice that frequently network graphs are sparse, meaning that $\hbox{den}(G_{obs}) = O\left(n^{-1}\right)$.  In such cases, we can then expect the performance of the naive estimator be dominated by the behavior of the Type I and II error rates $q$ and $p$, respectively.  Importantly, however, whether sparse or non-sparse, we note that there is no advantage in larger networks, i.e., as $n$ tends to infinity, if the error rates $p$ and $q$ are fixed relative to $n$, the accuracy in estimation of $g\left(W_{true}\right)$ in terms of its Lipschitz continuity is $O(1)$.  This observation motivates our construction of a estimator $\widehat{W}$ based on principles of denoising.

\subsection{Estimation through spectral denoising: The ideal estimator}
\label{3.2}

We now present the ideal estimator, i.e. in the case where we know the true eigenbasis $\{\psi_i\}_{i=1}^n$ of $W_{true}$, and characterize its performance.  We give several forms of the error bound, which vary with respect to what information about $W_{true}$ is assumed.

\begin{theorem}
\label{Ideal Estimator}
Let $\{\psi_i,\lambda_i\}_{i=1}^n$ be the orthonormal eigenvectors and associated eigenvalues of $W_{true}$ ordered decreasing according to $\{\lambda_i^2\}_{i=1}^n$.  Furthermore, define  the $n^2\times n^2$ covariance  matrix $C((x,y),(z,w)):=C(x,y,z,w)$ by

$$
\begin{aligned}
&C(x,y,z,w)= \mathbb{E}\left[ \left(W_{noise}(x,y)-q+(p+q)W_{true}(x,y) \right) \right.\\
&\hspace{2in} \cdot \left.\left(W_{noise}(z,w)-q+(p+q)W_{true}(z,w) \right)\right] \enskip ,
\end{aligned}$$
and define $$\widehat{W}_{ideal,s} = \sum_{j=1}^s \left<\psi_j,\widetilde{W}_{obs} \psi_j\right> \psi_j \psi_j^T.$$  Then,

\begin{enumerate}

\item (Spectral Moment Error Bound)

$$
\begin{aligned}
d\left(\widehat{W}_{ideal,s},W_{true}\right)^2 & = \frac{1}{(1-(p+q))^2} \sum_{j=1}^s \left<\psi_j\psi_j^T,C \psi_j\psi_j^T\right> + \sum_{j=s+1}^n \lambda_j^2\\
&\leq \frac{\sigma_{max} }{(1-(p+q))^2} \; s + \sum_{j=s+1}^n \lambda_j^2 \enskip ,\\
\end{aligned}
$$

where $\sigma_{max}$ is the spectral radius of $C$.  Furthermore, the minimum of the bound is given by the value of $s$ such that $$\lambda_{s+1}^2=\frac{\sigma_{max}}{(1-(p+q))^2},$$ if it exists; otherwise, the function is monotone decreasing and $s=n$.

\item (Degree Error Bound)  $$ \sum_{j=s+1}^n \lambda_j^2\leq \sum_{j=s+1}^n \bar{d}(j) $$ where $\{\bar{d}(i)\}_{i=1}^n$ is the degree sequence ordered descending.  Thus,

\begin{equation}
\label{Ideal Estimator MC} 
d\left(\widehat{W}_{ideal,s},W_{true}\right)^2  \leq \frac{\sigma_{max} }{(1-(p+q))^2}\enskip  s + \sum_{j=s+1}^n \bar{d}(j)
\end{equation}

where again, the minimum of the bound (if it exists) is attained at the value of $s$ for which $$\bar{d}(s+1)=\left\lfloor\frac{\sigma_{max}}{(1-(p+q))^2}\right\rfloor.$$

\item  (Spectral Radius for Independent Noise)  If $W_{noise}$ is independent across edges, $$\sigma_{max}=\max\{p(1-p),q(1-q)\}.$$

\item  (Asymptotic Relative Error for Power Law Networks)  Let $s_0$ be the index where equation (\ref{Ideal Estimator MC}) holds.  If $p,q\sim O(1)$, and $W_{true}$ has a power law degree distribution with exponent $\gamma>2$, then
\begin{equation}
\label{Ideal Estimator Power Law}
\frac{d\left(\widehat{W}_{ideal,s_0},W_{true}\right)^2}{d\left(\widetilde{W}_{obs},W_{true}\right)^2} \sim O\left(\frac{1}{n}\cdot \left( \frac{1}{\bar{d}(n)^{\gamma-2}} - \frac{\sigma_{max}}{ \bar{d}(1)^{\gamma-1}}\right) \right).
\end{equation}

In particular, if the noise is independent, this quantity is $O(1/n)$.

\end{enumerate}

\end{theorem}

The proof of these results are given in Appendix \ref{B.1}.  Intuitively, the theorem says that the projection of the noise onto each eigenspace of the true eigenbasis is approximately the same value $\sigma_{max}$, so that one can achieve a minimum mean squared error, provided the cost of projecting the noise onto the first $s$ modes offsets the bias incurred by ignoring the last $n-s$ modes.

For numerical simulations and validation of these results, see Section~\ref{4}.

\subsection{Estimation through spectral denoising: The empirical estimator}
\label{3.3}

The ideal estimator is useful in helping provide important insight into our estimation problem.  However, clearly it is of limited practical use, since we typically do not know the eigenfunctions of $W_{true}$.  If we instead try to use the empirical basis of $W_{obs}$, then (\ref{triangle estimate}) implies

\begin{equation}
\label{triangle inequality of relative error}
\frac{d\left(\widehat{W}_s,W_{true}\right)}{d\left(\widetilde{W}_{obs},W_{true}\right)} \leq \frac{d\left(\widehat{W}_s,\widehat{W}_{ideal,s}\right)}{d\left(\widetilde{W}_{obs},W_{true}\right)} + \frac{d\left(\widehat{W}_{ideal,s},W_{true}\right)}{d\left(\widetilde{W}_{obs},W_{true}\right)}.
\end{equation}

In the previous section, we bounded $d\left(\widehat{W}_{ideal,s},W_{true}\right)$.  In this section, we proceed to bound $d\left(\widehat{W}_s,\widehat{W}_{ideal,s}\right)$, and in the same spirit as the previous section, give several forms of the bound depending on what information is available.

\begin{theorem}
\label{Empirical Estimator}
Let $\{\phi_i,\mu_i\}_{i=1}^n$  be the orthonormal eigenvectors and associated eigenvalues of $\widetilde{W}_{obs}$, ordered decreasing according to $\{\mu_i^2\}_{i=1}^n$. Define
$$
\widehat{W}_s = \sum_{i=1}^s \left<\phi_i,\widetilde{W}_{obs}\phi_i\right> \phi_i\phi_i^T \enskip .
$$
Then,

\begin{enumerate}

\item  (Expected Sum of Spectral Moments \& Maximum Degree Bounds)

$$
d\left(\widehat{W}_s,\widehat{W}_{ideal,s}\right)^2 \leq 2\sum_{i=1}^s \mathbb{E}\left[\mu_i^2\right]\leq  2s\mathbb{E}\left[\widetilde{d}_{max}^2 \right]
$$

where $\widetilde{d}_{max}=\max_{i=1,\ldots,n} \{\widetilde{d}(i)\}$ and $\widetilde{d}(i)=\sum_{j=1}^n \widetilde{W}_{obs}(i,j)$.

\item  (Bound on the Expected Maximum Degree Squared for Independent Noise)  Let $\delta$ denote the maximum degree of $W_{true}$.  If the noise is independent, $$p+q<1,$$ and $$\log(n)\leq \delta(1-p)+q(n-1-\delta),$$

then

$$
\begin{aligned}
&\mathbb{E}\left[ \widetilde{d}_{max}^2\right] \leq \left( \delta+\frac{1+\sqrt{7}}{1-(p+q)} \sqrt{\log(n) \cdot [\delta(1-p)+q(n-1-\delta)]} \right)^2 \\
&\hspace{3in}+ \left[\frac{\max\{q,1-q\} }{1-(p+q)}\right]^2.
\end{aligned}
$$

\item  (Asymptotic Relative Error for Power Law Networks and Independent Noise)  
Combining the bound above with that of Theorem~2, and
applying (\ref{triangle inequality of relative error}),
the smallest resulting error bound for $d\left(\widehat{W}_s,W_{true}\right)$ is achieved if $s=1$.  
Furthermore, if $p,q\sim O(1)$, and $W_{true}$ has a power law degree distribution with exponent $\gamma>2$, then
\begin{equation}
\label{Empirical Estimator Power Law}
\begin{aligned}
\frac{d\left(\widehat{W}_1,W_{true}\right)^2}{d\left(\widetilde{W}_{obs},W_{true}\right)^2} \sim O\left( \left(\frac{\delta}{n}\right)^2 + \frac{\log(n)}{n}  \right).
\end{aligned}
\end{equation}

In particular, if $$\delta\sim o\left(\sqrt{n\log(n)}\right),$$ this becomes $O\left( \log(n)/ n\right).$

\end{enumerate}
\end{theorem}

The proof of this theorem is given in Appendix \ref{B.2}.  

A few things are worth pointing out in this theorem:

\begin{enumerate}

\item  (Concentration of Degree and Noise)  The first part holds in full generality, while the second part holds for independent noise only.  However, one can straightforwardly extend these latter results to general noise models.  The key ingredient to bounding $\mathbb{E}\left[\widetilde{d}_{max}^2\right]$ in any extension is a concentration inequality for the degree $i$ of $\widetilde{W}_{obs}$ to the degree $i$ of $W_{true}$.  In our work, we have used the concentration inequality for sums of independent Bernoulli random variables found in \cite{ChungandLu}.

\item  (Comparison of the Ideal and Empirical Estimators)  When $p,q\sim O(1)$, $W_{true}$ has a power law degree distribution with exponent $\gamma>2$, and in addition, the noise is independent, then equations (\ref{Ideal Estimator Power Law}) and (\ref{Empirical Estimator Power Law}) imply
$$
\frac{d\left(\widehat{W}_{ideal,s_0},W_{true}\right)^2}{d\left(\widetilde{W}_{obs},W_{true}\right)^2} \sim O\left(\frac{1}{n}\right)
$$
and
$$
\begin{aligned}
\frac{d\left(\widehat{W}_1,W_{true}\right)^2}{d\left(\widetilde{W}_{obs},W_{true}\right)^2} \sim O\left( \left(\frac{\delta}{n}\right)^2 + \frac{\log(n)}{n}  \right) \enskip ,
\end{aligned}
$$
where $\delta$ is the maximum degree of $W_{true}$.  Combining this with equation (\ref{statistical estimation}), we have with respect to estimation of the statistic $g$, that
$$
\begin{aligned}
\frac{\left\{\mathbb{E} \left[ g\left(\widehat{W}_{ideal,s_0}\right) - 
		g\left(W_{true}\right) \right]^2 \right\}^{\frac{1}{2}}}{n^2}
    &\sim O\left(\frac{1}{n}\right)
 \end{aligned}
$$

and

$$
\begin{aligned}
\frac{\left\{\mathbb{E} \left[ g\left(\widehat{W}_{1}\right) - 
		g\left(W_{true}\right) \right]^2 \right\}^{\frac{1}{2}}}{n^2}
  &\sim O\left( \left(\frac{\delta}{n}\right)^2 + \frac{\log(n)}{n}  \right)
\enskip .
  \end{aligned}
$$

We see that in the case of independent noise,  the optimal rate for the ideal estimator in this case is $1/n$.  In the empirical estimator, we find that if $\delta\sim o\left(\sqrt{n\log(n)}\right),$ then the rate becomes $\log(n)/n$ and $(\delta/n)^2$ otherwise.  

The dependence on $\delta$ is not surprising since the bound depends on the expected maximum degree squared.  The presence of the factor $\log(n)/n$ is due to the concentration inequality for the independent noise and is related to the window in the concentration in which tail probabilities scale like $1/n^2$.  It is because of this reason that we are forced to choose $s=1$.  For different types of dependencies, one can expect this to change.

\item    It is interesting to note that in interpreting Theorem~\ref{Empirical Estimator}, the expected Frobenius norm squared between the projection of a matrix onto its first $s$-dimensional largest eigenspaces and its projection onto the first $s$-dimensional largest eigenspaces of its expected value is dominated by the $s^{th}$ partial second spectral moment.  To date, the only estimate to be found on this quantity is in the form of the extreme partial trace formula of \cite{Tao}, but more useful bounds akin to Theorem \ref{Ideal Estimator}, where one can bound this quantity by, say, the more interpretable first few dominant degrees, are unknown.  

One can expect to obtain a more refined bound akin to this partial second spectral moment bound that includes the angles between the true eigenvectors of $W_{true}$ and the empirical ones of $\widetilde{W}_{obs}$, so that in the limit as $p,q\rightarrow 0$, the error in approximating the true eigenbasis by the empirical one tends to zero.  For our purposes, however, since $p,q\sim O(1)$ and $n$ is growing, the dominant term in the error is those of the expected square eigenvalues which we find in the bound.  In improving the bound by including eigenvectors angles, one would find $s:=s(p,q)$ for any kind of condition on $p,q$, however we leave such improvements for future work.

We should note, however, that in the case of the ideal estimator $W_{ideal,s}$, we do have $s:=s(p,q,n)$ since by Theorem \ref{Ideal Estimator}, the smallest error occurs when $$\lambda_{s+1}^2 =\frac{\sigma_{max}}{(1-(p+q))^2}.$$

\end{enumerate}

For numerical simulations and validation of these results, see Section \ref{4}.

\subsection{Continuity of network summary statistics}
\label{3.4}

The results of the previous sections establish that it is possible to estimate $g\left(W_{true}\right)$ substantially better when the argument to $g$, rather than being simply the observation $W_{obs}$ (as is common in practice), is instead an appropriately denoised version of $W_{obs}$, i.e., $\widehat{W}_{s}$.  There are many such network summary statistics $g$ used in applications.  As mentioned earlier, however, our results apply only to those that are sufficiently smooth in $W$.  While we have not attempted to provide an exhaustive survey of the numerous (and still growing) list of network summary statistics available, the following result characterizes a representative collection.

\begin{theorem}
\label{contstatpf}
Consider the class of adjacency matrices $W$ for all connected, undirected (possibly weighted) graphs $G$.  
\begin{enumerate}

\item  The following network summary statistics are not continuous functions of the adjacency matrix in any norm:

\begin{enumerate}
\item  Geodesic distances;

\item  Betweenness centrality;

\item  Closeness centrality.
\end{enumerate}

\item  The following network summary statistics are continuous functions of $W$ in any norm:
\begin{enumerate}

\item  Degree centrality;  

\item  The number of $k$ paths between vertices;

\item  Conductance of a set $S\subset V$ on the set of graphs for which $S$ contains at least one edge with weight $\delta>0$;

\item Eigenvector Centrality in the metric $$d(v_1,v_2)=\sqrt{1-\left| \left<v_1,v_2\right> \right|^2};$$

\item  Density.
\end{enumerate}
\end{enumerate}

In particular in (2), (a), (c) and (e) are Lipschitz while $(b)$ and $(d)$ are locally Lipschitz.
\end{theorem}

The proof of these results is given in Appendix \ref{C}.  Note that for the three summary statistics where continuity fails, all are functions of shortest paths.  The geodesic distance between pairs of nodes is the length of the (not necessarily unique) shortest path from one to the other.  Betweenness centrality is a vertex-specific quantity, which effectively counts the number of shortest paths between vertex pairs that pass through a given vertex.  Finally, closeness centrality is simply the inverse of the total distance of a vertex to all others in the graph.  In all cases, essentially, continuity fails due to the sensitivity of shortest paths to perturbations.

In contrast, the summary statistics for which continuity holds are representative of a number of different types of quantities.  Degree centrality is in fact just vertex degree, while density was defined earlier, as $m/{n\choose 2}$, which is proportional to the average vertex degree over the entire graph.  Hence, our results on estimation pertain to  degree, a fundamental quantity in graph theory and network analysis, on both local and global scales.  On the other hand, the number of $k$-paths between vertices and the conductance of a set $S$ are both relevant to the study of flows in networks.  Finally, eigenvector centrality of a given vertex refers to the corresponding entry of the eigenvector of $W$ associated with the largest eigenvalue.  It is intimately related to the stationary probabilities of a random walk on the graph $G$, and is a measure of vertex `importance' that, among other uses, plays a role in the ranking of results from a Google search.

\section{Numerical results}
\label{4}

In this section we present three sets of numerical illustrations of our work.  Section \ref{4.1} addresses the case where one can use the ideal estimator $\widehat{W}_{ideal,s}$ in circumstances where one knows the true eigenbasis of $W_{true}$.  Although artificial, from a practical point of view, these results both provide additional insight into the theory developed in Section~\ref{3.2} for the ideal estimator and establish a useful baseline.  In Sections~\ref{4.2} and~\ref{4.3} we then give examples using real-world data sets.  The protein-protein interaction network of~\cite{PPI} (gleaned from BioGRID) is used in Section~\ref{4.2} to illustrate the performance that might be expected of our estimator when knowledge of the Type I and II error probabilities $q$ and $p$ are taken from the experimental literature \citep{Traver}.  In particular, using simulation with a range of realistic error values, we demonstrate a strong robustness of the superiority of our proposed estimator $\widehat{W}_s$ over the observed $W_{obs}$ to misspecification of the true $q$ and $p$.  In contrast, a gene regulatory network for the pheromone response pathway of \cite{Roberts:2000uq}, constructed using gene expression data of~\cite{Brem2005}, is used in Section~\ref{4.3} to illustrate the application of our proposed method in the situation where there is no other recourse but to estimate $q$ and $p$ from the same data used to infer the network $G_{obs}$ (and hence $W_{obs}$) itself.


\subsection{Evaluation of $\widehat{W}_{ideal,s}$ on simulated data}
\label{4.1}

Motivated by the success in using $\widehat{W}_{ideal,s}$ over the raw data $\widetilde{W}_{obs}$ in Section \ref{3.2}, we confine ourselves to this choice of the optimal basis and ask how well we can denoise the observation $\widetilde{W}_{obs}.$  Note that this choice represents an ideal situation, in which we already know a basis defined explicitly in terms of the object we seek to recover -- but we do not know the values of the coefficients to assign to the elements of this basis.  As such, our results below should be understood as providing insight into the best we might hope to do in our estimation problem.  We begin by providing examples of constructions of networks from simpler ones where this information is known.

Let $G_i $ denote a family of graphs on the vertices $V_i$ with $|V_i|=n_i$ and adjacency matrices $W_i$.  There exist very general Cartesian type operations  \citep{cvet1} on these graphs to obtain graphs on the vertex set $$\mathfrak{V}=V_1\times \cdots \times V_n.$$ 

Let $\mathcal{B}\subseteq \{0,1\}^n-\{(0,\ldots,0)\}=H_0^n$.  The non-complete extended p-sum, (NEPS) of $G_1,\ldots,G_n$, $\mathfrak{G}_{\mathcal{B}}$, on $\mathfrak{V}$ is defined by,

$$
\begin{aligned}
(x_1,\ldots,x_n)\sim (y_1,\ldots,y_n) & \; {\rm if \; and \; only\; if \; there\; exists}\; \beta \in \mathcal{B} \; {\rm such \; that\;} x_i=y_i \\
& {\rm whenever\;} \beta_i=0 \; {\rm and} \; { x_i\sim y_i} \; {\rm whenever} \; \beta_i=1
\end{aligned}
$$

Intuitively, the nontrivial elements of the hypercube $\mathcal{B}$ indicate in which components two  points $x,y\in \mathcal{V}$ are joined.  Of particular interest are:

\begin{enumerate}
\item  The Cartesian product, or sum, $G_1+\cdots+G_n$ with $\mathcal{B}$ consisting of the standard basis vectors in $\mathbb{R}^n$.  This guarantees that for each $n$-tuple, if we travel in each direction $e_j$ (fixing all vertices with the same component except $j$), we trace out $G_j$.

By taking $G_i=P_k$, a path graph on $k$ vertices, this include $k$-lattices in $\mathbb{R}^n$.  By taking $G_i=C_{k-1}$, the cycle on $k-1$ vertices, this also includes toroidal lattices.

\item  The tensor product, or product, $G_1\otimes \cdots \otimes G_n$ with $\mathcal{B}=\{(1,\ldots,1)\}$.  In contrast to the Cartesian product, this guarantees that an n-tuple $x\in \mathcal{V}$ is connected to all other n-tuples provided the corresponding component vertices are connecting in their graphs.

\item  The strong product, $G_1\ast \cdots \ast G_n$ with $\mathcal{B}=H_0^n$, which is a union of the two previous graphs.
\end{enumerate}

We have the following results \citep{cvet1}

\begin{prop}
\label{svetadjprop}
The NEPS $\mathfrak{G}_{\mathcal{B}}$ has adjacency matrix $$\mathfrak{W}=\sum_{\beta \in \mathcal{B}} W_1^{\beta_1}\otimes \cdots \otimes W_n^{\beta_n}\enskip ,$$ where $W_k^0$ is the identity matrix of the same size as $W_k$ and $W_k^1=W_k$.
\end{prop}

\begin{prop}
\label{cartlap}
Using the same notation as in proposition \ref{svetadjprop}, let $\mathfrak{G}_{\mathcal{B}}$ be a NEPS graph with 

$$\mathfrak{W}=\sum_{\beta \in \mathcal{B}} W_1^{\beta_1}\otimes \cdots \otimes W_n^{\beta_n}.$$

If $\lambda_{i1},\ldots,\lambda_{ik_i}$ are the eigenvalues of $G_i$ for $i=1,\ldots, n$, then the spectrum of the NEPS of $G_1,\ldots, G_n$ with basis $\mathcal{B}$ consists of all possible values $$\Lambda_{i_1,\cdots,i_n}=\sum_{\beta\in \mathcal{B}} \lambda_{1i_1}^{\beta_1}\cdots \lambda_{ni_n}^{\beta_n}, \; \; \; i_h=1,\ldots, k_n; \; \; \; h=1,\ldots, n.$$

If $x_{i,j}$, $j=1,\ldots, k_i$ are the linearly independent eigenvectors of $G_i$ with $W_i x_{i,j}=\lambda_{ij} x_{i,j}$, then $$x=x_{1i_1}\otimes \cdots \otimes x_{ni_n}$$ is an eigenvector of $\mathfrak{W}$ with eigenvalue $\Lambda_{i_1,\ldots,i_n}$.
\end{prop}

We now present applications of these results relevant to when we can assume knowledge of the true eigenfunctions of an underlying graph.

\begin{enumerate}
\item  For the toroidal $k$-lattice in $\mathbb{R}^n$, $T^{k}_n=\sum_{j=1}^n C_{k-1}$.  For $C_{k-1}$, note that each vertex has degree 2.  Note that $W_{C_{k-1}}=P+P^{-1}$ where $P$ is a permutation matrix determined by a cyclic permutation of length $k-1$.   If $\omega$ is a $k-1$ root of unity, then $$x_{\omega}=(1,\omega,\omega^2,\ldots,\omega^{k-2})^T$$ is an eigenvector of $P$ with eigenvalue $\omega$.  From this, we immediately see that the eigenvalues of $W_{C_{k-1}}$ are $$\lambda_j=2\cos\left(\frac{2\pi j}{k-1}\right)  \; \; j=1,\ldots,k-1$$ with eigenvectors 

$$(y_j)(\ell)=C_j  \cos\left(\frac{2\pi j \ell}{k-1}\right)\; \; j=1,\ldots,k-1$$ where $C_j$ is a normalization constant.  We now apply proposition \ref{cartlap} for the eigenvectors and eigenvalues of the toroidal lattice.  One should note that most of the eigenvalues have multiplicity 2, so that the eigenspaces are actually two-dimensional.

\item  For a $k$-lattice in $\mathbb{R}^n$, $L^k_n=\sum_{j=1}^n P_k$.  From \citep{mieghem}, the eigenvalues of $P_k$ are given by 

$$\lambda_j=2\cos\left(\frac{\pi}{k+1}j\right), \; \; \; j=1,\ldots,k.$$

The nontrivial eigenvectors are given by

$$(x_m)_{\ell} = C_m \sin\left( \frac{\pi m \ell }{k+1}\right), \; \; \; 1\leq \ell\leq k $$  where $C_m$ is a normalization constant.  We can now apply proposition \ref{cartlap} for the eigenvectors and eigenvalues of the lattice.

\end{enumerate}

Using these results, we consider the simple case of a genus five toroidal lattice obtained from an undirected 5 cycle, $\sum_{j=1}^5 C_5$, using independent noise and $p=0.3$, $q=0.4$.  This graph has $n=3,125$ vertices with $m=15,625$ edges.

\begin{figure}[htp!]
\includegraphics[width=12cm, height=8cm]{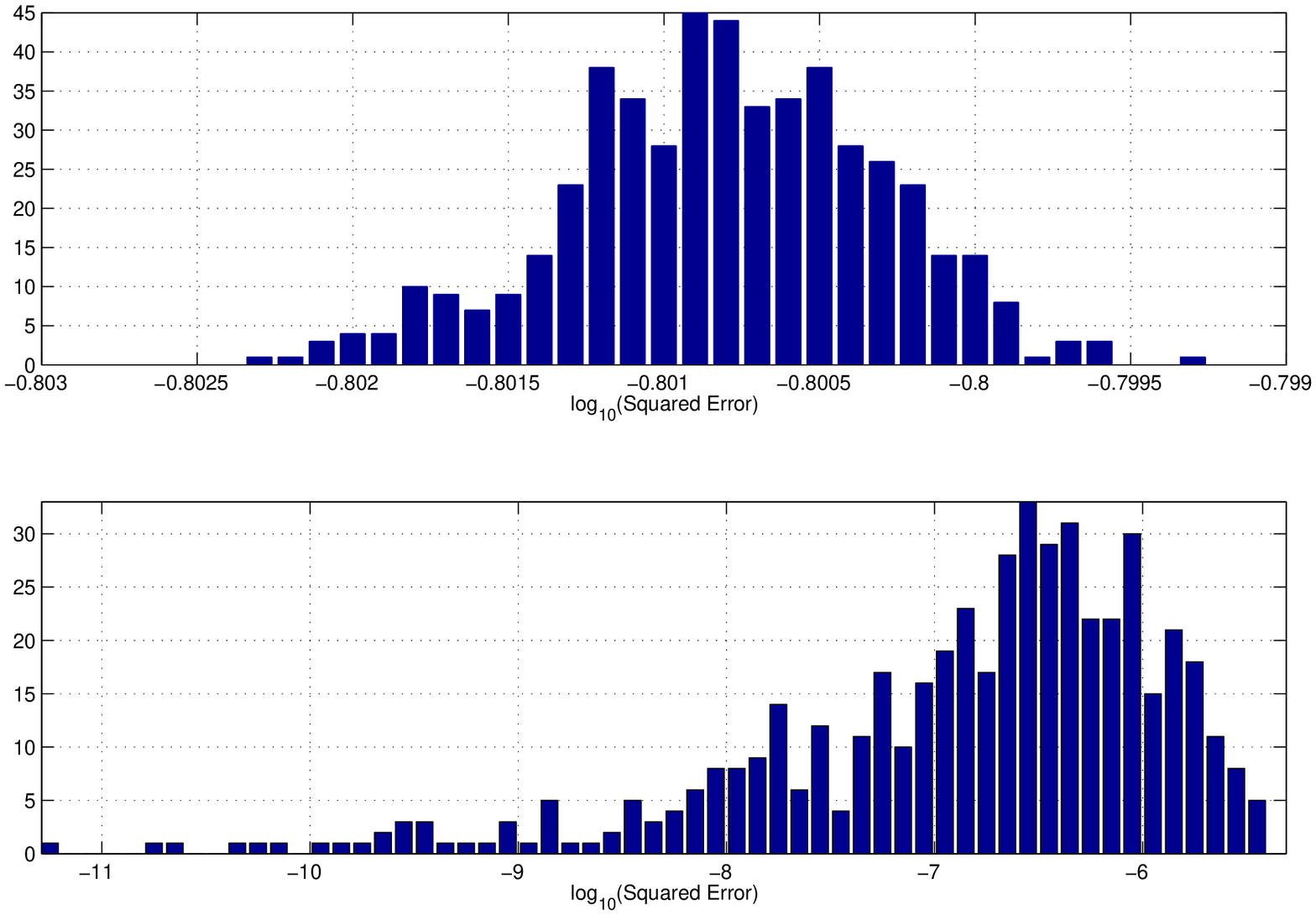}
\caption{Histograms of $\log_{10}$-squared deviations of densities for the genus five toroidal lattice using an undirected five cycle.  Top: histogram of $\log_{10}$-squared deviations of the densities for $W_{obs}$ to $W_{true}$.  Bottom: histogram of $\log_{10}$-squared deviations of the densities for $\widehat{W}_{ideal,1,845}$ to $W_{true}$.  Simulations are over $N=500$ samples with $p=0.3$ and $q=0.4$ using independent noise.}
\label{5x5TDensity}
\end{figure}

In Figure \ref{5x5TDensity} we plot the histogram of the $\log_{10}$-squared deviation of the density of edges in the network using $W_{obs}$ from that of $W_{true}$ obtained over 500 samples in the top panel.  In the bottom panel, we plot a histogram of the  $\log_{10}$-squared deviation of the density of edges using our estimator $\widehat{W}_{ideal,1,845}$ from that of $W_{true}$ for the same 100 samples.   As we will see in the next figure, the relative error for $\widehat{W}_{ideal,s}$ is minimized at $s=1,845$ which motivates this choice, and we choose to plot the  $\log_{10}$-squared errors for visual clarity.  
We note that our estimator has a mean-squared error of $5.83 \times 10^{-7}$ with standard deviation $8.218\times 10^{-7}$ in measuring the true density while $W_{obs}$ has a mean-squared error of $1.582\times 10^{-1}$ with standard deviation $1.825\times 10^{-4}$ in measuring the true density.   In other words, our estimator yields, on average, an improvement of six orders of magnitude over a naive estimator that simply uses the observed network.

\begin{figure}[htp!]
\includegraphics[width=12cm, height=10cm]{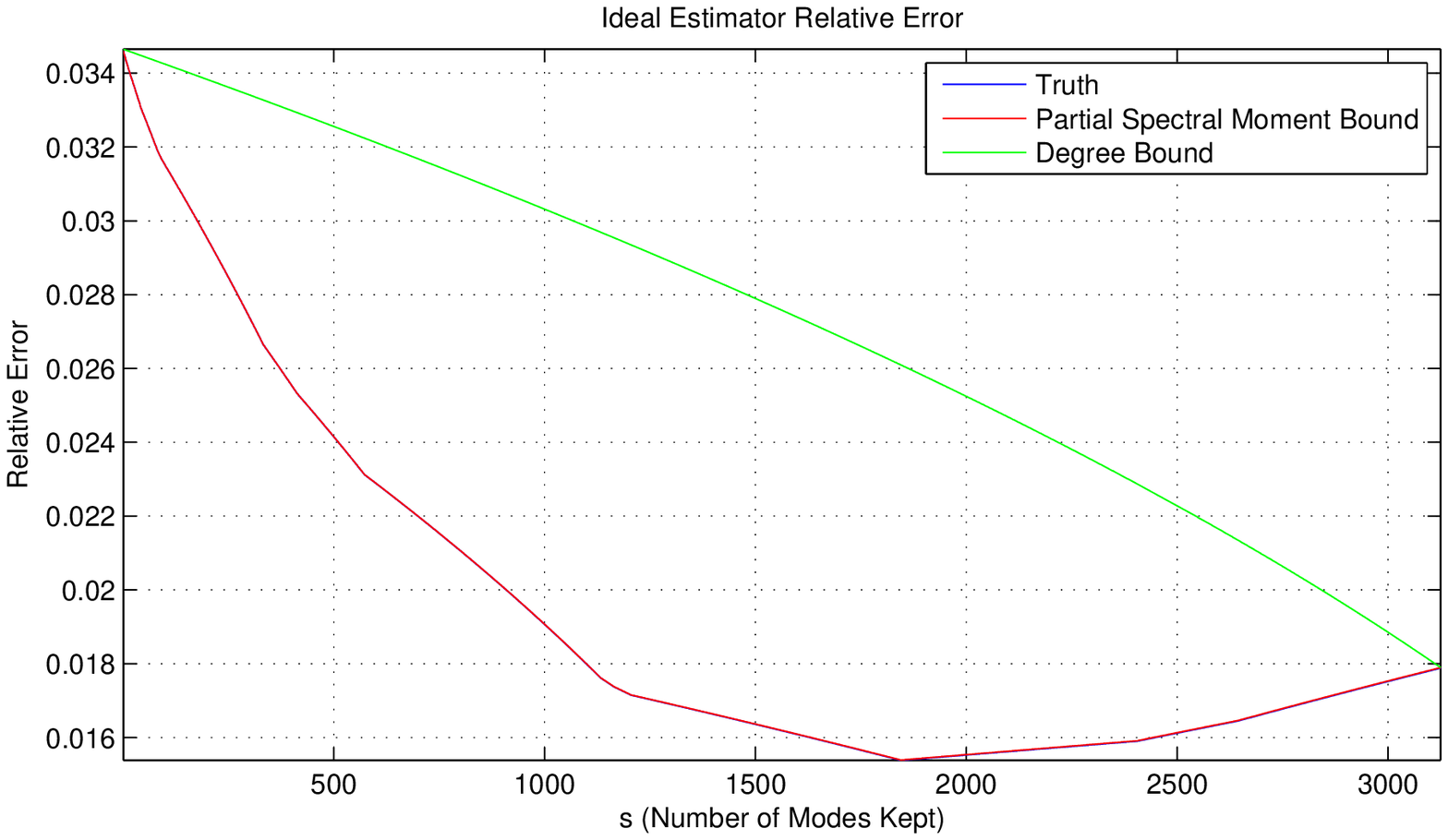}
\caption{Plot of the relative errors bounds on $d\left(\widehat{W}_{ideal,s},W_{true}\right)/d\left(\widetilde{W}_{obs},W_{true}\right)$ from theorems \ref{Ideal Estimator} and  \ref{Ideal Estimator}  for $W_{true}$ taken from a genus 5 toroidal lattice using an undirected five-cycle using $p=0.3$ and $q=0.4$ with independent noise.  We note that the blue curve is underneath the red curve, indicating the estimate on the true value of the relative error is very precise.}
\label{5x5TRE}
\end{figure}

In Figure \ref{5x5TRE}, we plot the relative error curves from Section \ref{three}, Theorem \ref{Ideal Estimator} for $W_{ideal,s}$ in this data set using the eigenvectors given above for $\sum_{j=1}^5 C_5$ across all $3,125$ modes.  We note that when using the ideal estimator, the true relative error and the relative error through the spectral bound are nearly identical, and achieve a minimum relative error at $s=1,845$ of approximately $0.0154$.  The degree bound on the relative error, however, while more interpretable, is worse, with an approximate error of $0.027$, since it linearly interpolates between the relative error at $s=1$ and $s=3,125$ since $\sum_{j=1}^5 C_5$ has constant degree $10$ everywhere.

\subsection{Application of $\widehat{W}_{s}$ to protein-protein interaction networks}
\label{4.2}

Here we present an illustration of our method using a version of the yeast protein-protein interaction (PPI) network, as extracted from BioGRID and analyzed in~\cite{PPI}.  The declaration of `edge' or `non-edge' status in such networks is typically based on experiments aimed at assessing the affinity of protein pairs for binding to each other.  Such measurement of interactions among proteins is widely recognized to be extremely noisy.  \cite{Traver} recently have summarized the fairly substantial literature on quantifying the Type I and Type II error probabilities $q$ and $p$ that can be expected in practice for yeast PPI under standard experimental regimes.  We use this network and the error rates drawn from this literature in a quasi-simulation study that will illustrate the robustness that can be expected of our method to misspecification of $q$ and $p$, since such misspecification is nearly certain in practice.  (We present a full application to real data in the next section, in the context of gene regulatory networks.)

The original PPI network has $n=5,151$ vertices and $m=31,201$ edges.
In their Table~1, \cite{Traver} summarize (average) false-positive rates
found in the literature.  Using what appears to be their preferred value,
we set $q=0.35$.  Similarly, in their Table~2 is found a summary of various
projections of the overall network size (i.e., number of edges) in the true 
yeast interactome.  Given their mean projected size of $52,500$ edges, and
the observed size of $31,201$ in our own network, we are led to use a
false-negative rate of $p=0.40$ (i.e., $(52,500-31,201)/52,500=0.4057$).

We then simulated noisy versions of the protein-protein interaction network of~\cite{PPI} in the same manner as described in Section~\ref{4.1}, using the values $q=0.35$ and $p=0.40$.  Equipped with knowledge of these true values of $q$ and $p$, we find that our estimator $\widehat{W}_1$ has an average mean-squared error of $5.516 \times 10^{-6}$ with standard deviation $8.8\times 10^{-8}$ in measuring the true density, while $W_{obs}$ has an average mean-squared error of $1.213\times 10^{-1}$ with standard deviation $9.54\times 10^{-5}$.  That is, our estimator yields, on average, an improvement of roughly five orders of magnitude.

Motivated by the success of $\widehat{W}_1$ over $W_{observed}$ for the true values of $p=0.4$ and $q=0.35$, we may ask how robust it is with respect to these parameters since they will not be known precisely in practice.  Rather, an experimentalist typically will be more comfortable stating that these parameters lie within a certain range of values.  In Table \ref{PPI W1 Summary}, therefore, we give the mean-squared error and the standard deviation of squared errors in estimating the density using $\widehat{W}_1$, for values of $p$ and $q$ that deviate from their true values by $\pm 0.05$, using the same $500$ samples for each choice.  Examing the results, we see that performance seems to be driven most strongly by the accuracy to which we know the Type I error probability $q$, and that when this value is known accurately, we maintain our five orders of magnitude improvement.  This result is encouraging, because the Type I rate will likely be more laboratory-dependent and therefore should be known with some reasonable accuracy.  On the other hand, even when we misspecify $q$ by as much as $\pm 0.05$ (i.e, more than $10\%$), we generally still observe an order of magnitude improvement, except in the case when both $q$ and $p$ are specified $0.05$ too high.
\begin{table}[htp!]\footnotesize
  \caption{Mean-squared Errors and Standard Deviation Summaries for $\widehat{W}_{1}$ for the Protein-Protein Interaction Network of \cite{PPI} from BioGRID for different values of Misspecified Type I and Type II Errors using 500 Samples}
  \begin{tabular}{| c | c | c | c | c |}
    \hline & $q=0.3$ & $q=0.35$ & q=0.4\\
      \hline  $p=0.35$ & $(2.02\times 10^{-2}, 1\times10^{-4})$ & $(5.519\times10^{-6}, 7.3\times10^{-8})$ & $(4\times 10^{-2}, 2\times 10^{-4} )$ \\
\hline  $p=0.4$&$(2.76\times 10^{-2}, 2\times 10^{-4})$ & $(5.516\times 10^{-6},8.8\times 10^{-8})$ & $(6.22\times 10^{-2}, 3\times 10^{-4})$\\
\hline  $p=0.45$&$(4\times 10^{-2}, 2\times 10^{-4})$ & $(5.512\times 10^{-6}, 1.09\times 10^{-7})$ & $(1.1\times 10^{-1}, 6\times 10^{-4})$\\ \hline
  \end{tabular}
  \label{PPI W1 Summary}
\end{table}

\subsection{Application of $\widehat{W}_{s}$ to gene regulatory networks}
\label{4.3}

In this section, we consider the fully realistic scenario where one is given only one observation of $W_{observed}$ of the network, and, moreover, the Type I and Type II error rates $q$ and $p$ must be obtained from the same data used to create the network itself.

\begin{figure}[htp!]
\includegraphics[width=12cm, height=10cm]{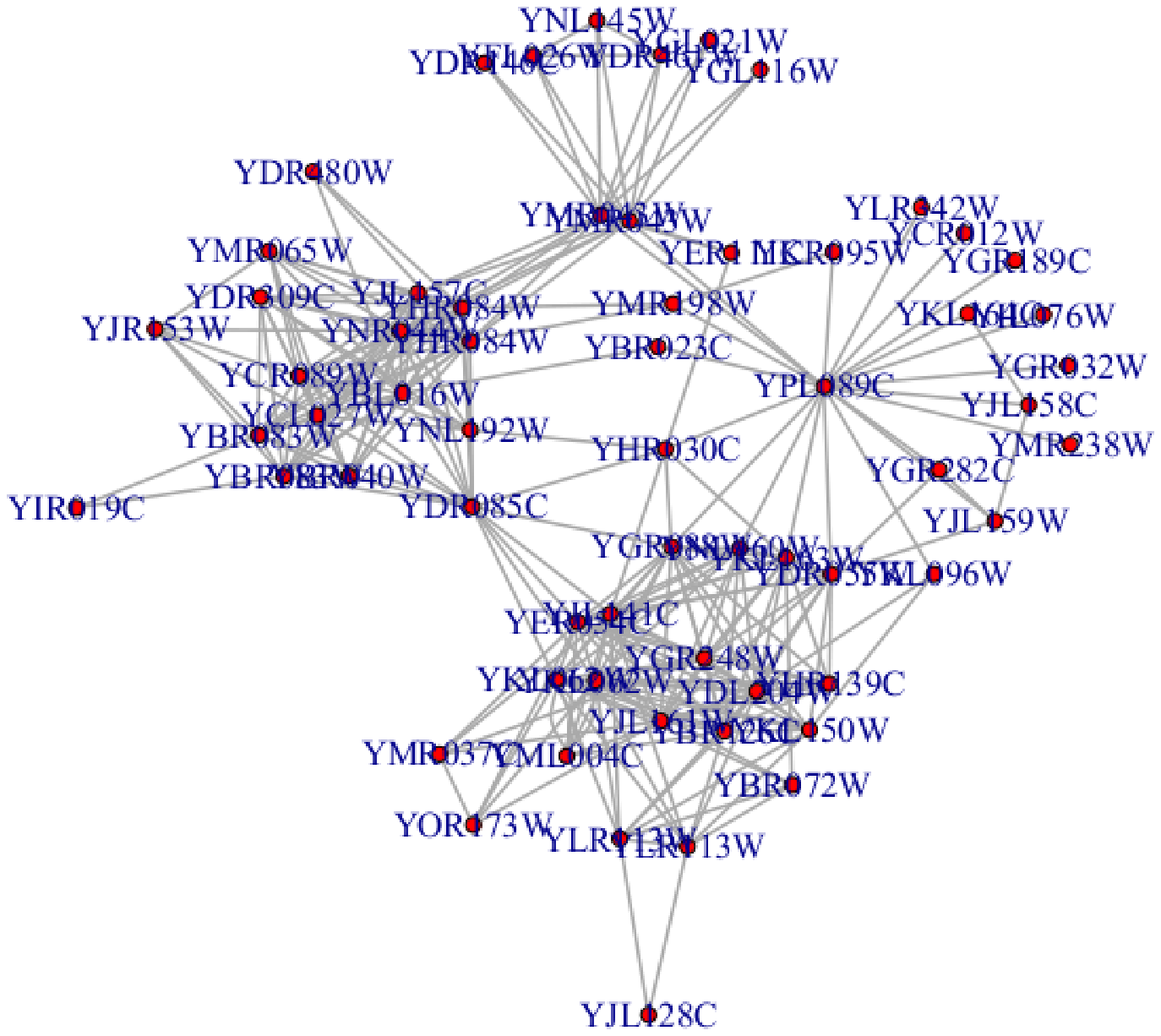}
\caption{A visualization of the gene regulatory network described in the text.}
\label{PPI Graph Viz}
\end{figure}

The network we use here, i.e., $W_{observed}$, is built from the gene expression data of \cite{Brem:2005kx} on a set of genes from the pheromone response pathway, as selected by~\cite{Roberts:2000uq}.  A visualization of this network is shown in Figure~\ref{PPI Graph Viz}.  There are $63$ nodes and $554$ edges in this network.  It is an example of an association network, in that the edges are meant to indicate that there is a nontrivial association between the expression levels of the incident genes.  Such association networks, in turn, are felt to be indicative of gene regulatory relationships.  

We used a testing-based approach to constructing this network, wherein sufficiently `large' values of a test statistic are taken as evidence against a null hypothesis that there is no edge.  (See \cite[Ch 7.3]{kola:2009} for a general summary of this and other such methods for inference of association networks.)  The test statistic used here is simply the empirical Pearson moment correlation of expression between pairs of genes over the set of experiments.  Following standard practice in this area, the threshold for declaring significance of a given test statistic value (i.e., for declaring an edge in the network) was chosen so as to control for multiple testing.  A common framework in which to execute such control is through false discovery rate principles.  While there are many approaches to doing so, we find it convenient here to employ empirical null principles \citep{efron2010large}, as implemented in the {\em fdrtool} package of~\cite{strimmer2008fdrtool}.  In particular, {\em fdrtool} assumes a two-component mixture model for the marginal distribution of the empirical correlations across all pairs of nodes, and produces estimates of the distribution under the null and alternative, as well as the mixing parameter.  Applying this approach to our data, using the default settings, and the specification of `correlation' as the statistic in calling {\em fdrtool}, and accepting the threshold resulting from the software, we obtained estimates of $p=0.008886004$ and $q=0.00407613$.

Given this observed network, and using these estimates of $q$ and $p$ in computing $\widehat{W}_1$, we then calculated, by way of illustration, the corresponding estimates of the summary statistics of density, eigenvector centrality, and degree centrality.  We summarize these estimates in Table \ref{GW1 Summary Statistics}.  For density, we report the values obtained by using $W_{observed}$ and $\widehat{W}_1$.  However, since eigenvector centrality and degree centrality are vector quantities, we report, for a given centrality measure, just a single-measure summary, the so-called centralization \citep{Freeman}.  The latter is a measure of how central is the most central node of a network in relation to how central all the other nodes are. That is, if $C_x\left(v\right)$ is any centrality measure $x$ evaluated at the node $v$, and $C_x\left(v^{\ast}\right)$ is the largest such measure in the network, then the centralization of the network is given by 
$$C_x=\frac{\sum_{v=1}^{n} \left[C_x\left(v^{\ast}\right)-C_x\left(v\right)\right]}{\max_{{\rm all\; graphs \; with\; n\; nodes}} \sum_{v=1}^{n} \left[C_x\left(v^{\ast}\right)-C_x\left(v\right)\right]} \enskip .$$
We note that in the definition of centralization, the denominator is maximized when considering a single disconnected node, and has value $n$ in this case.

In summarizing the results from Table~\ref{GW1 Summary Statistics}, we find that the estimates of the density decreases but that the centralizations of both eigenvector and degree centrality stay approximately the same. (We note, however, that the estimates of individual eigenvector and degree centralities differ for the two estimators (not shown).)
\begin{table}[htp!]\footnotesize
  \caption{Summary statistics for the gene regulatory network of Figure~\ref{PPI Graph Viz}.}
    \begin{tabular}{| c | c |c|}
    \hline & $W_{observed}$ & $\widehat{W}_1$\\
      \hline  Density & $0.1418$&$0.1095$  \\
\hline  Eigenvector Centrality& $0.0862$&$0.0860$ \\ 
\hline  Degree Centrality&$12.2063$&$13.5619$ \\ \hline
  \end{tabular}
  \label{GW1 Summary Statistics}
\end{table}

\section{Concluding Remarks}
\label{5}

While it is now common across the various scientific disciplines to construct network-based representations of complex systems and to then summarize those representations with various statistics (e.g. density, betweenness, conductance, etc.) there has to date been little to no acknowledgement of the necessary propagation of error from network construction to network summary.  Effectively, in simply reporting the statistics of their ``noisy" networks, investigators are using plug-in estimators.  

In this paper, we formalize this problem and lay an initial formulation for its solution.  Specifically, we have constructed several spectral-based estimators of network adjacency matrices in order to more accurately estimate Lipschitz continuous network summary statistics in mean square error

$$
\frac{\left\{\mathbb{E} \left[ g\left(W_1\right) - 
		g\left(W_2\right) \right]^2 \right\}^{\frac{1}{2}}}{n^2}
  \le  C  \frac{ d\left(W_1,W_2\right)}{n}.\enskip
$$
where $d\left(W_1,W_2\right)=   \left\{\mathbb{E}\left[ ||W_1 - W_2 ||_F^2\right] \right\}^{\frac{1}{2}} /n$.

Under the additive noise model $W_{obs}=W_{true}+W_{noise}$ that has $W_{true}$ fixed and Type I and II errors the same across edges, we first construct an unbiased estimator, $\widetilde{W}_{obs}$ that can be computed directly from the observed data, and use spectral projection onto its first $s$ dominant modes in absolute value to obtain our empirical estimator, $\widehat{W}_s$, treating $s$ as a model selection parameter that controls a bias/variance tradeoff.  We bound the performance of $\widehat{W}_s$ to $W_{true}$ by comparing it through the performance of another estimator, the oracle estimator $\widehat{W}_{ideal,s}$, in which the unbiased estimator is projected onto the eigenbasis of the true network.  By bounding the performance of both the oracle and empirical estimator, we show how to choose $s$, and obtain asymptotic results on the relative error, $d\left(\widehat{W},W_{true}\right)/d\left(\widetilde{W}_{obs},W_{true}\right)$.
We conclude by giving several network summary statistics that are and are not continuous with respect to any norm, and give numerical examples employing $\widehat{W}_{ideal,s}$ and $\widehat{W}_s$, to both synthetic and real-world data.

Our work is significant in laying a critical new foundation stone in the general area of network inference.  At the same time, it should be remembered that our theoretical work makes the important assumption that the Type I and II error probabilities (i.e., $q$ and $p$) are known.  While this assumption, analogous to claiming knowledge of the noise level in traditional regression modeling, is undoubtedly false in practice, it is particularly useful in allowing us to work with centered versions of the network adjacency matrix which, in turn, allows for a noticeably greater level of interpretation of the various results we produce.  Furthermore, we have demonstrated, through our numerical work that not only does knowledge of exact Type I and II error rates yield substantial improvements over naive plug-estimation (i.e., by multiple orders of magnitude), but in addition, these improvements appear to be rather robust to misspecification (with our examples involving misspecification by more than $10\%$ at particularly high error rates).  Finally, we have shown that it is possible, using techniques and tools standard to the inference of association networks in computational biology, to estimate the Type I and II error rates.  

An important next step, building upon our work here, would be to develop an analogous set of theoretical bounds on the risk of estimators like ours that explicitly take into account the uncertainty due to any estimation of $q$ and $p$.  We note, however, that to date the development of methods for network inference (which spans the literatures of numerous fields) far outstrips the formal assessment of the corresponding theoretical properties, i.e., including, in particular, a formal characterization of resulting Type I and Type II error rates.  For example, even in the context of regression-based methods of network inference based on principles of $\ell_1$ penalization, which are recently popular in the statistics literature, even the accurate estimation of $p$-values (which could, for instance, be used as input to something like {\em fdrtool} in a manner similar to our analysis in Section~\ref{4.3}) is still an open challenge, with methods like that of~\cite{liu2012high} beginning to make some early progress.

Our work is also significant in using the spectral theory of both the eigenvectors and eigenvalues of $W$ to solve real-world problems.  To the best knowledge of the authors, eigenvectors of $W$ for most classes of graphs of practical interest are poorly understood to date.  This is the first work in which these bases are given an applied interpretation in statistical estimation of network summary statistics.  Furthermore, it also shows that the so-called eigenvector centrality vector (the dominant eigenfunction of a network) not only can be used to rank vertices (as is done), but contains substantially more information when combined with its corresponding eigenvalue; indeed, we see this in our work using the empirical estimator where the minimal relative error is achieved with $s=1$.  The histograms of density, for example, illustrate that by simply projecting the observed signal onto this centrality vector, several orders of magnitude of improved accuracy can be achieved.  

Some additional, technical comments are in order for future extensions of this work.  First, it is straightforward to remove the assumption of the same Type I and Type II errors ($p$ and $q$) across all edges, at the cost of increasing all relative error bounds.  However, one would still require knowledge of the number of edges since this controls how many Type II errors are being committed.  In this spirit, it is the belief of the authors that much of this can be extended straightforwardly to directed networks using SVD in lieu of an eigendecomposition.

Second, we have shown that $\widehat{W}_{ideal,s}$ is substantially better than $\widehat{W}_s$ in estimating $W_{true}$ and hence all Lipschitz continuous statistics.  This raises the question as to whether one can estimate the eigenbasis of $W_{true}$ better than just using the eigenbasis of $\widetilde{W}_{obs}$.  For example, if one uses a spanning set of vectors that still parametrize the eigenspaces that are linearly dependent, redundancy can make projection onto certain subspaces more stable and hence might make estimation of the eigenspaces more tenable.  Further considerations are so-called star bases in \citep{cvet1}.

Third, when bounding the relative error of $\widehat{W}_s$ to $\widehat{W}_{ideal,s}$, we neglected the analysis of the angles between the eigenspaces of $\widetilde{W}_{obs}$, and $W_{true}$ since in the limit of large $n$ the dominating effect is in the eigenvalues.  Indeed, if one were to consider the limit of $p,q\rightarrow 0$ for fixed $n$, then these angles must be taken into account, and as a result, $s:=s(p,q)$.  In the general case then, we can expect $s:=s(p,q,n)$ and a more careful analysis might reduce all relative error bounds, even in the case we have been considering, $p,q\sim O(1)$ and $n\rightarrow \infty$.  Such an analysis, however, requires a detailed understanding of the concentration of the angles $\cos^2\left(\phi_i,\psi_j\right)$ where $\{\phi_i\}_{i=1}^n$ are the eigenbasis of $\widetilde{W}_{obs}$ and $\{\psi_i\}_{i=1}^n$ are the eigenbasis of $W_{true}$ both sorted descending according to their respective squared eigenvalues.    

Fourth, and perhaps most interesting, would be a further detailed analysis of the covariance matrix $C$ appearing in theorem \ref{Ideal Estimator} for dependencies in the noise across edges and its spectral properties, since most noise in many real-world applications are in fact dependent.  Indeed, at the time of this writing, dependencies in additive noise has been poorly understood in the context of $C$ and this object has appeared nowhere explicitly in the literature.  Once the nature of this dependency is understood, it would also allow for an extension of the relative error of $\widehat{W}_{ideal,s}$ to $\widehat{W}_s$ and would yield a more nontrivial bias/variance tradeoff than simply taking $s=1$ in the empirical estimator.

\bibliographystyle{plainnat}

\appendix

\section{Proof of Theorem 1}
\label{A}

\begin{enumerate}
\item  In the original basis, 
$$
\begin{aligned}
&\mathbb{E}\left[\widetilde{W}_{obs}(i,j)-W_{true}(i,j)\right]\\
&=\mathbb{E}\left[\frac{W_{obs}(i,j)-qW_{K_n}(i,j)}{1-(p+q)}-W_{true}(i,j)\right]\\
&=\frac{1}{1-(p+q)}\mathbb{E}\left[ W_{noise}(i,j)-q+(p+q)W_{true}(i,j)\right]\\
&=0.
\end{aligned}
$$

Changing bases using a deterministic base change proves the result, as the corresponding entries are linear combinations of mean zero random variables.

\item  By the previous result,

$$
\begin{aligned}
\mathbb{E}\left[||\widetilde{W}_{obs}-W_{true}||_F^2\right]&=\sum_{x,y=1}^n \mathbb{E}\left[\ \left|\widetilde{W}_{obs}(x,y)-W_{true}(x,y)\right|^2\right]\\
&=\frac{p(1-p)2m+q(1-q)\left[n(n-1)-2m\right]}{(1-(p+q))^2}\\
&=\frac{2\left(p(1-p)m+q(1-q)\left[{n\choose 2}-m\right]\right)}{(1-(p+q))^2}.
\end{aligned}
$$

The inequality (\ref{eq:mse.g.naive}) follows in turn from the Lipschitz property,
$$\left| g(W_1)-g(W_2)\right|\leq C||W_1-W_2||_1\leq C\cdot n \cdot ||W_1-W_2||_F,$$ since 
$$\frac{|g(W)-g(W_0)|}{n^2} \leq C\frac{||W-W_0||_F}{n}\enskip .$$
\end{enumerate}

\begin{flushright}
$\square$
\end{flushright}

\section{Proofs of Theorems 2 \& 3}
\label{B}
\subsection{Proof of Theorem 2}
\label{B.1}

\begin{enumerate}

\item  Using that

$$\widehat{W}_{ideal,s}= \sum_{j=1}^s \left<\psi_j,\widetilde{W}_{obs}\psi_j\right> \psi_j \psi_j^T$$ and $$W_{true}=\sum_{j=1}^n \left<\psi_j,W_{true}\psi_j\right> \psi_j\psi_j^T$$ we collect the first $s$ terms and use orthonormality of $\{\psi_j\}_{j=1}^n$ to obtain,

$$
\begin{aligned}
&\mathbb{E}\left[ ||\widehat{W}_{ideal,s} - W_{true}||_F^2\right] \\
&= \mathbb{E}\left[ \left| \left| \sum_{j=1}^s \left<\psi_j,\widetilde{W}_{obs}\psi_j\right> \psi_j \psi_j^T - \sum_{j=1}^n \left<\psi_j,W_{true}\psi_j\right> \psi_j\psi_j^T\right| \right|_F^2\right]\\
&= \mathbb{E}\left[ \left| \left| \sum_{j=1}^s \left<\psi_j,\left(\widetilde{W}_{obs}-W_{true}\right)\psi_j\right> \psi_j \psi_j^T - \sum_{j=s+1}^n \left<\psi_j,W_{true}\psi_j\right> \psi_j\psi_j^T\right| \right|_F^2\right]\\
&=\sum_{j=1}^s \mathbb{E}\left[ \left| \left<\psi_j, (\widetilde{W}_{obs}-W_{true}) \psi_j\right>\right|^2\right]+\sum_{j=s+1}^n \lambda_j^2\\
&=\frac{1}{(1-(p+q))^2}\sum_{j=1}^s \mathbb{E}\left[ \left|\left<\psi_j,(W_{noise}-qW_{K_n}+(p+q)W_{true})\psi_j\right>\right|^2\right]\\
&\hspace{2.5in}+\sum_{j=s+1}^n \lambda_j^2\\
\end{aligned}
$$

where in the last equality, we have used the definition of $$\widetilde{W}_{obs}=\frac{W_{obs}-qW_{K_n}}{1-(p+q)}.$$
Now, 
$$\begin{aligned}
&\frac{1}{(1-(p+q))^2}\sum_{j=1}^s \mathbb{E}\left[ \left|\left<\psi_j,(W_{noise}-qW_{K_n}+(p+q)W_{true})\psi_j\right>\right|^2\right]\\
&=\frac{1}{(1-(p+q))^2} \sum_{j=1}^s \mathbb{E}\left[ \left| \sum_{x,y=1}^n \left(W_{noise}(x,y)-q \right. \right. \right. \\
& \hspace{2in}\left. \left. \left.+(p+q)W_{true}(x,y) \right)\psi_j(x)\psi_j(y)\right|^2\right]\\
&=\frac{ \sum_{j=1}^s \sum_{x,y,z,w=1}^n C(x,y,z,w)\psi_j(x)\psi_j(y)\psi_j(z)\psi_j(w) }{(1-(p+q))^2}\\
&=\frac{ \sum_{j=1}^s \left<\psi_j\psi_j^T,C \psi_j\psi_j^T\right>}{(1-(p+q))^2} \\
&\leq \frac{\sigma_{max}}{(1-(p+q))^2} s\\
\end{aligned}
$$

where in the last line, $\sigma_{max}$ is the spectral radius of $C$.  Thus,

$$\begin{aligned}
&\mathbb{E}\left[ ||\widehat{W}_{ideal,s} - W_{true}||_F^2\right] \leq \frac{\sigma_{max}}{(1-(p+q))^2} s + \sum_{j=s+1}^n \lambda_j^2.\\
\end{aligned}
$$

If we call the bound $f(s)$, notice that

$$f(s+1)-f(s) = \frac{\sigma_{max}}{(1-(p+q))^2} -\lambda_{s+1}^2.$$  Since $\{\lambda_k^2\}_{k=1}^n$ are ordered descending, it is clear that the minimum value will occur when $\lambda_{s+1}^2 = {\sigma_{max}}/{(1-(p+q))^2} $.

\item  We use the following result from \citep{mieghem}.

\begin{prop}
Let $A$ be a real symmetric $n\times n$ matrix with eigenvalues  $$\lambda_n(A)\leq \cdots \leq \lambda_1(A)$$ and ordered diagonal elements $$d(n)\leq \cdots \leq d(1).$$  Then, for any $1\leq k \leq n$, it holds that $$\sum_{j=1}^k d(j)\leq \sum_{j=1}^k \lambda_j(A).$$
\end{prop}

Now, consider the matrix $W^2$ with diagonal entries $W^2(i,i)=d(i)$.  By invariance of the trace, $\sum_{j=1}^n \lambda^2_j =\sum_{j=1}^n d(j)=2|E|=2m$, so that applying the above proposition,

$$\sum_{j=k+1}^n \bar{d}(j) = 2E - \sum_{j=1}^k \bar{d}(j)\geq 2E-\sum_{j=1}^k \lambda^2_j = \sum_{j=k+1}^n \lambda_j^2.$$

The result now follows immediately from the bound in (1).  The statement about the minimum follows the same argument as that in (1).

\item  If $W_{noise}$ is independent across edges, the definition of $C$ implies that $C$ is diagonal, with diagonal entries either $p(1-p)$ or $q(1-q)$.  Thus, $\sigma_{max}=\max\{p(1-p),q(1-q)\}.$

\item  Using theorem \ref{Raw Data}, and the above results (2)

\begin{equation}
\label{ideal comp}
\begin{aligned}
&\frac{d\left(\widehat{W}_{ideal,s},W_{true}\right)^2}{d\left(\widetilde{W}_{obs},W_{true}\right)^2} \\
& \leq 
 \left[ \frac{s\left(\frac{\sigma_{max}}{(1-(p+q))^2}\right)+\sum_{j=s+1}^n \bar{d}(j)}{d\left(\widetilde{W}_{obs},W_{true}\right)^2}\right]\\
 &=\frac{(1-(p+q))^2}{2} \left[ \frac{s\left(\frac{\sigma_{max}}{(1-(p+q))^2}\right)+\sum_{j=s+1}^n \bar{d}(j)}{p(1-p)m+q(1-q)\left[{n\choose 2}-m\right]}\right]\\
\end{aligned}
\end{equation}

the minimization condition

$$\left\lfloor \frac{\sigma_{max}}{(1-(p+q))^2}\right\rfloor = \bar{d}(s+1)$$

implies that this occurs precisely when we choose $s$ to be the value when the $(s+1)^{st}$ largest degree grows as fast as $\left\lfloor \sigma_{max} /(1-(p+q))^2 \right\rfloor$, implying that $s$ is the number of vertices in the network whose degree grow faster than $\left\lfloor \frac{\sigma_{max}}{(1-(p+q))^2}\right\rfloor$.  We take $s$ to be the first index, $s_0$, for which this occurs.

If $W_{true}$ has a power-law degree distribution, i.e. $P(x)=\pi/x^{\gamma}$ with $\gamma>2$ where $P(x)$ is the fraction of vertices in $W_{true}$ with degree $x$, the fraction of vertices whose degree is larger than $\bar{d}(s_0+1)$ is given by 

$$
\begin{aligned}
\sum_{\stackrel{\bar{d}(s_0+1)<k\leq \bar{d}(1)}{k\; \; degree\; \; of \; \; W_{true}}} P(k)&=\sum_{\stackrel{\bar{d}(s_0+1) <k\leq\bar{d}(1)}{{k\; \; degree\; \; of \; \; W_{true}}}} \frac{\pi}{k^{\gamma}} \\
&\leq \pi\int_{d(s_0+1) }^{\bar{d}(1)} \frac{dx}{x^{\gamma}}\\
&=\frac{\pi}{\gamma-1} \cdot \left[\frac{1}{\bar{d}(s_0+1)^{\gamma-1}}-\frac{1}{\bar{d}(1)^{\gamma-1}}\right]
\end{aligned}
$$

so that

$$s_0=n\cdot \sum_{ \bar{d}(s_0+1)<k\leq \bar{d}(1)} P(k)\leq \frac{\pi n}{\gamma-1} \cdot \left[\frac{1}{\bar{d}(s_0+1)^{\gamma-1}}-\frac{1}{\bar{d}(1)^{\gamma-1}}\right]$$

Similarly,

$$
\begin{aligned}
\frac{1}{n}\sum_{j=s_0+1}^n \bar{d}(j) &= \sum_{\bar{d}(n)\leq k\leq  \bar{d}(s_0+1)} P(k) k\\
&=\pi\sum_{\bar{d}(n)\leq k\leq  \bar{d}(s_0+1)} \frac{1}{k^{\gamma-1}} \\
&\leq \pi\left[ \frac{1}{\bar{d}(n)^{\gamma-1}} +\frac{1}{\gamma-2} \left( \frac{1}{\bar{d}(n)^{\gamma-2}} - \frac{1}{\bar{d}(s_0+1)^{\gamma-2}}\right) \right]
\end{aligned}
$$

so that 

$$\sum_{j=s_0+1}^n \bar{d}(j) \leq \pi n \left[ \frac{1}{\bar{d}(n)^{\gamma-1}}+\frac{1}{\gamma-2} \left( \frac{1}{\bar{d}(n)^{\gamma-2}} - \frac{1}{\bar{d}(s_0+1)^{\gamma-2}}\right) \right]$$

Substituting these into equation \ref{ideal comp},

\begin{equation}
\label{ideal comp 2}
\begin{aligned}
&\frac{d\left(\widehat{W}_{ideal,s_0},W_{true}\right)^2}{d\left(\widetilde{W}_{obs},W_{true}\right)^2}\\
&\leq  \left[ \frac{s_0\left(\frac{\sigma_{max}}{(1-(p+q))^2}\right)+\sum_{j=s_0+1}^n \bar{d}(j)}{d\left(\widetilde{W}_{obs},W_{true}\right)^2}\right]\\
 &=\frac{\frac{(1-(p+q))^2}{2} \cdot n\pi}{p(1-p)m+q(1-q)\left[{n\choose 2}-m\right]}\\
 & \hspace{.5in}\cdot  \left[ \frac{1}{\gamma-1} \cdot \left[\frac{1}{\bar{d}(s_0+1)^{\gamma-1}}-\frac{1}{\bar{d}(1)^{\gamma-1}}\right]\left(\frac{\sigma_{max}}{(1-(p+q))^2}\right)\right.\\
 &\hspace{1in}\left.+\frac{1}{\bar{d}(n)^{\gamma-1}}+\frac{1}{\gamma-2} \left( \frac{1}{\bar{d}(n)^{\gamma-2}} - \frac{1}{\bar{d}(s_0+1)^{\gamma-2}}\right) \right]\\
 &\leq \frac{\frac{(1-(p+q))^2}{2} \cdot \frac{n\pi}{\gamma-2}}{p(1-p)m+q(1-q)\left[{n\choose 2}-m\right]} \\
 &\hspace{1in}\cdot  \left[ \frac{3}{\bar{d}(n)^{\gamma-2}} - \frac{\sigma_{max}}{(1-(p+q))^2 \bar{d}(1)^{\gamma-1}} \right] \\
 &\sim O\left(\frac{1}{n}\cdot \left( \frac{1}{\bar{d}(n)^{\gamma-2}} - \frac{\sigma_{max}}{ \bar{d}(1)^{\gamma-1}}\right) \right)\\
\end{aligned}
\end{equation}

when $p,q\sim O(1)$.

If the noise if independent, $\sigma_{max}\sim O(1)$ since $p,q\sim O(1)$, so that the asymptotic becomes $O\left({1}/{n}\right)$.

\end{enumerate}

\begin{flushright}
$\square$
\end{flushright}

\subsection{Proof of Theorem 3}
\label{B.2}

\begin{enumerate}

\item  

For notational convenience, set $b_k=\left<\psi_k,\widetilde{W}_{obs}\psi_k\right>$.  Then,

$$
\begin{aligned}
&||\sum_{n=1}^s \mu_k \phi_k\phi_k^T - \sum_{k=1}^sb_k \psi_k \psi_k^T||_F^2\\
&=\sum_{x,y=1}^n \left| \sum_{n=1}^s \mu_k \phi_k(x)\phi_k(y) - \sum_{k=1}^sb_k \psi_k(x) \psi_k(y)\right|^2\\
&=\sum_{k=1}^s \left[\mu_k^2+b_k^2-2\sum_{k,\ell}\mu_k b_{\ell}\left<\phi_k,\psi_{\ell}\right>^2\right]\\
\end{aligned}
$$
Grouping the last two terms together and using

 $$b_k=\left<\psi_k,\widetilde{W}_{obs}\psi_k\right>=\sum_{\ell=1}^n \mu_{\ell} \cos^2\left(\phi_{\ell},\psi_k\right) \enskip ,$$
 
$$
\begin{aligned}
&\sum_{k=1}^s \left[\mu_k^2+b_k^2-2\sum_{k,\ell}\mu_k b_{\ell}\left<\phi_k,\psi_{\ell}\right>^2\right]\\
&=\sum_{\ell=1}^s \mu_{\ell}^2 + \sum_{\ell=1}^s b_{\ell}\left[\sum_{k=s+1}^n \mu_k \cos^2\left(\phi_k,\psi_{\ell}\right) - \sum_{k=1}^s \mu_k \cos^2 \left(\psi_{\ell},\phi_k\right)\right]\\
&=\sum_{\ell=1}^s \mu_{\ell}^2 +\sum_{\ell=1}^s \left\{\left(\sum_{k=s+1}^n \mu_k\cos^2\left(\phi_k,\psi_{\ell}\right)\right)^2 \right.\\
&\hspace{2in}\left.-\left(\sum_{k=1}^s \mu_k\cos^2\left(\phi_k,\psi_{\ell}\right)\right)^2 \right\} \\
\end{aligned}
$$

Dropping the third term and bounding the middle by pushing $\mu_k\rightarrow \mu_{s+1}$,

$$
\begin{aligned}
&=\sum_{\ell=1}^s \mu_{\ell}^2 +\sum_{\ell=1}^s \left\{\left(\sum_{k=s+1}^n \mu_k\cos^2\left(\phi_k,\psi_{\ell}\right)\right)^2 \right.\\
&\hspace{2in}\left.-\left(\sum_{k=1}^s \mu_k\cos^2\left(\phi_k,\psi_{\ell}\right)\right)^2 \right\} \\
&\leq \sum_{\ell=1}^s \mu_{\ell}^2 +\sum_{\ell=1}^s\mu_{s+1}^2  \left(\sum_{k=s+1}^n \cos^2\left(\phi_k,\psi_{\ell}\right)\right)^2\\
&\leq \sum_{\ell=1}^s \mu_{\ell}^2 + \sum_{\ell=1}^s \mu_{\ell}^2 \cdot 1\\
&=2\sum_{\ell=1}^s \mu_{\ell}^2.
\end{aligned}
$$

where in the penultimate inequality, we have used that $\{\mu_i^2\}_{i=1}^n$ is monotone decreasing and $\sum_{k=1}^n \cos^2\left(\phi_k\psi_{\ell}\right) = ||\psi_{\ell}||^2=1$.  Thus,

$$||\sum_{n=1}^s \mu_k \phi_k\phi_k^T - \sum_{k=1}^sb_k \psi_k \psi_k^T||_F^2 \leq 2\sum_{\ell=1}^s \mu_{\ell}^2$$ so that taking expected values yields,

$$d\left(\widehat{W}_s,\widehat{W}_{ideal,s}\right)^2 \leq 2\sum_{\ell=1}^s \mathbb{E}\left[\mu_{\ell}^2\right].$$

Since there are no self loops, Gerschgorin's theorem \citep{mieghem} implies for $i=1,\ldots,n$,

$$-\widetilde{d}_{max}\leq \mu_i \leq \widetilde{d}_{max}$$ 

so that

$$d\left(\widehat{W}_s,\widehat{W}_{ideal,s}\right)^2 \leq 2\sum_{\ell=1}^s \mathbb{E}\left[\mu_{\ell}^2\right]\leq 2s\mathbb{E}\left[\widetilde{d}_{max}^2\right].$$

\item  We prove the bound on $\mathbb{E}\left[\widetilde{d}_{max}^2\right]$ since the final statement is a direct result of it and (1) above.  First, we need the following lemma on a concentration inequality for the maximum degree.

\begin{lemma}
\label{maximum degree probability bound}
Let $\delta$ denote the maximum degree of $W_{true}$ and $\widetilde{d}_{max}$ denote the maximum column sum for $\widetilde{W}_{obs}$.  Define

$$\epsilon=\frac{1+\sqrt{7}}{1-(p+q)} \sqrt{\log(n) \cdot [\delta(1-p)+q(n-1-\delta)]}$$

and suppose that the noise is independent across edges.  

If $$p+q<1$$ and $$\log(n)\leq \delta(1-p)+q(n-1-\delta)$$ then 

$$\mathbb{P}\left[\widetilde{d}_{max}>\delta+\epsilon \right] \leq \frac{1}{n^2}.$$
\end{lemma}
{\bf Proof of Lemma \ref{maximum degree  probability bound}:}

Let $\delta$ denote the maximum degree of $W_{true}$ with individual degrees $d(i)$, and $\widetilde{d}_{max}$ denote the maximum column sum for $\widetilde{W}_{obs}$ with individual column sums $\widetilde{d}(i)$.  Then for any $\epsilon>0$,

$$
\begin{aligned}
&\mathbb{P}\left[\widetilde{d}_{max}-\delta>+\epsilon\right]\\
&=\mathbb{P}\left[\widetilde{d}_{max}>\delta+\epsilon\right]\\
&=\mathbb{P}\left[ \bigcup_{i=1}^n \left\{ \widetilde{d}(i)>\delta+\epsilon\right\}\right]\\
&\leq \sum_{i=1}^n \mathbb{P}\left[\widetilde{d}(i)>\delta+\epsilon\right]\\
&\leq \sum_{i=1}^n \mathbb{P}\left[\widetilde{d}(i)>d(i)+\epsilon\right]\\
&= \sum_{i=1}^n \mathbb{P}\left[ \sum_{j=1}^n W_{obs}(i,j)>d(i)(1-p)\right.\\
&\hspace{2in}\left.+q(n-1-d(i))+\epsilon(1-(p+q))\right]
\end{aligned}
$$

$$\begin{aligned}
&\leq \sum_{i=1}^n \exp\left[-\frac{1}{2} \frac{\epsilon^2(1-(p+q))^2}{d(i)(1-p)+q(n-1-d(i))+\frac{\epsilon(1-(p+q))}{3}}\right]\\
&= \sum_{i=1}^n \exp\left[-\frac{1}{2} \frac{\epsilon^2(1-(p+q))^2}{d(i)(1-(p+q))+q(n-1)+\frac{\epsilon(1-(p+q))}{3}}\right]\\
&\leq  n \exp\left[-\frac{1}{2} \frac{\epsilon^2(1-(p+q))^2}{\delta(1-(p+q))+q(n-1)+\frac{\epsilon(1-(p+q))}{3}}\right]\\
&=  n \exp\left[-\frac{1}{2}\cdot \frac{1}{\delta(1-(p+q))+q(n-1) }\cdot \frac{\epsilon^2(1-(p+q))^2}{1+\frac{\epsilon(1-(p+q))}{3\left[\delta(1-(p+q))+q(n-1)\right]}}\right]\\
\end{aligned}
$$

where the last inequality follows because $p+q<1$, and in the penultimate inequality we have used the concentration inequality from \cite{ChungandLu}.

With the choice of $$\epsilon=\frac{1+\sqrt{7}}{1-(p+q)} \sqrt{\log(n) \cdot [\delta(1-p)+q(n-1-\delta)]}$$ and since

$$\log(n)\leq \delta(1-(p+q))+q(n-1)=\delta(1-p)+q(n-1-\delta)$$

this implies

$$
\begin{aligned}
\mathbb{P}\left[\widetilde{d}_{max}-\delta>+\epsilon\right]&\leq n \exp\left[-\frac{1}{2} \cdot \frac{(1+\sqrt{7})^2 \log(n)}{1+ \frac{1+\sqrt{7}}{3} \sqrt{\frac{\log(n)}{\delta(1-(p+q))+q(n-1)}}}\right]\\
&\leq  n \exp\left[-\frac{1}{2} \cdot \frac{(1+\sqrt{7})^2 \log(n)}{1+ \frac{1+\sqrt{7}}{3}}\right]\\
&= n \exp\left[-3\log(n)\right]\\
&=\frac{1}{n^2}
\end{aligned}
$$

\begin{flushright}
$\triangle$
\end{flushright}

\begin{corollary}
\label{maximum degree mean squared bound}
In the same notation and conditions as theorem \ref{maximum degree  probability bound},
$$
\begin{aligned}
&\mathbb{E}\left[ \widetilde{d}_{max}^2\right] \leq \left( \delta+\frac{1+\sqrt{7}}{1-(p+q)} \sqrt{\log(n) \cdot [\delta(1-p)+q(n-1-\delta)]} \right)^2 \\
&\hspace{2in}+ \left[\frac{\max\{q,1-q\} }{1-(p+q)}\right]^2.
\end{aligned}
$$
\end{corollary}

{\bf Proof of Corollary \ref{maximum degree mean squared bound}:}

By theorem \ref{maximum degree  probability bound},

$$
\begin{aligned}
\mathbb{E}\left[\widetilde{d}_{max}^2\right]&= \mathbb{E}\left[\widetilde{d}_{max}^2 \Big| \widetilde{d}_{max}\leq\delta+\epsilon\right]\mathbb{P}\left[\widetilde{d}_{max}\leq \delta+\epsilon\right]\\
&\hspace{1in}+\mathbb{E}\left[\widetilde{d}_{max}^2 \Big| \widetilde{d}_{max}>\delta+\epsilon\right]\mathbb{P}\left[\widetilde{d}_{max}<\delta+\epsilon\right]\\
&\leq \left(\delta+\epsilon\right)^2 + \mathbb{E}\left[\widetilde{d}_{max}^2 \Big| \widetilde{d}_{max}>\delta+\epsilon\right]\cdot \frac{1}{n^2}
\end{aligned}
$$

To complete the proof, note that $\widetilde{W}_{obs}=\frac{W_{obs}-qW_{K_n}}{1-(p+q)}$ implies,

$$-\frac{q(n-1)}{1-(p+q)} \leq \frac{d(i)-q(n-1)}{1-(p+q)} = \widetilde{d}(i)\leq \frac{(1-q)(n-1)}{1-(p+q)}$$

so that

$$-\frac{q(n-1)}{1-(p+q)} \leq  \widetilde{d}_{max}\leq \frac{(1-q)(n-1)}{1-(p+q)}$$

and $\widetilde{d}_{max}^2 \leq \left[\frac{\max\{q,1-q\}(n-1)}{1-(p+q)}\right]^2.$

\begin{flushright}
$\triangle$
\end{flushright}

\item  To gain insight into the asymptotic nature of $d\left(\widehat{W}_s,W_{true}\right)$ we first replace the triangle inequality in equation \ref{triangle inequality of relative error},

 $$\frac{d\left(\widehat{W}_s,W_{true}\right)}{d\left(\widetilde{W}_{obs},W_{true}\right)} \leq \frac{d\left(\widehat{W}_s,\widehat{W}_{ideal,s}\right)}{d\left(\widetilde{W}_{obs},W_{true}\right)} + \frac{d\left(\widehat{W}_{ideal,s},W_{true}\right)}{d\left(\widetilde{W}_{obs},W_{true}\right)}$$

 with the statement of convexity 
 
 $$\frac{d\left(\widehat{W}_s,W_{true}\right)^2}{d\left(\widetilde{W}_{obs},W_{true}\right)^2} \leq 2\left[ \frac{d\left(\widehat{W}_s,\widehat{W}_{ideal,s}\right)^2}{d\left(\widetilde{W}_{obs},W_{true}\right)^2} + \frac{d\left(\widehat{W}_{ideal,s},W_{true}\right)^2}{d\left(\widetilde{W}_{obs},W_{true}\right)^2}\right]$$
 
 to obtain

\begin{equation}
\begin{aligned}
\label{empirical comp}
\frac{d\left(\widehat{W}_s,W_{true}\right)^2}{d\left(\widetilde{W}_{obs},W_{true}\right)^2} &\leq 2\left[ \frac{d\left(\widehat{W}_s,\widehat{W}_{ideal,s}\right)^2}{d\left(\widetilde{W}_{obs},W_{true}\right)^2} + \frac{d\left(\widehat{W}_{ideal,s},W_{true}\right)^2}{d\left(\widetilde{W}_{obs},W_{true}\right)^2}\right]\\
& \leq 2\left[ \frac{s\left(\frac{\sigma_{max}}{(1-(p+q))^2}+2\mathbb{E}\left[\widetilde{d}_{max}^2\right]\right)+\sum_{j=s+1}^n \bar{d}(j)}{d\left(\widetilde{W}_{obs},W_{true}\right)^2}\right]
\end{aligned}
\end{equation}

where we have used convexity to add the two error bounds in theorems \ref{Ideal Estimator} and \ref{Empirical Estimator} and $\bar{d}$ is the ordered decreasing degree sequence of $W_{true}$.  Minimizing over $s$ by finding a critical point, we find that the first order difference is given by

$$\left(\frac{\sigma_{max}}{(1-(p+q))^2} + \left(\delta+\sqrt{n\log{n}}\right)^2\right) - \bar{d}(s+1)$$

Note that this difference is always positive so that the bound in equation \ref{empirical comp} is always monotone increasing and we are forced to choose $s=1$ as the argmin.  Since $p,q\sim O(1)$,

$$
\begin{aligned}
\label{empiricalcomp}
\frac{d\left(\widehat{W}_s,W_{true}\right)^2}{d\left(\widetilde{W}_{obs},W_{true}\right)^2} \sim O\left(  \frac{\frac{\sigma_{max}}{(1-(p+q))^2}+\left(\delta+\sqrt{n\log{n}}\right)^2+\sum_{j=2}^n \bar{d}(j)}{n^2}\right)
\end{aligned}
$$

Since we are in the regime of independent noise, and $p,q\sim O(1)$, we have from theorem \ref{Ideal Estimator} $\sigma_{max}\sim O(1)$ and so $\sigma_{max}/[n^2(1-(p+q))^2] \sim O(1/n^2).$  If $W_{true}$ has a power law degree distribution with $P(k)=\pi/k^{\gamma}$ with $\gamma>2$ equal to the fraction of vertices with degree $k$, 

$$
\begin{aligned}
\frac{1}{n}\sum_{j=2}^n \bar{d}(j) &=  \pi \sum_{\stackrel{\bar{d}(n)\leq k \leq \bar{d}(2)}{k \; \; degree \; \; of \; \; W_{true}}} P(k)k \\
& = \pi \sum_{\stackrel{\bar{d}(n)\leq k \leq \bar{d}(2)}{k \; \; degree \; \; of \; \; W_{true}}} \frac{1}{k^{\gamma-1}}\\
&\leq \pi\left[\frac{1}{\bar{d}(n)^{\gamma-1}}+\frac{1}{\gamma-2} \left(\frac{1}{\bar{d}(n)^{\gamma-2}}-\frac{1}{\bar{d}(2)^{\gamma-2}}\right)\right]\\
\end{aligned}
$$

where in the last inequality we have used an integral bound after isolating the $k=\bar{d}(n)$ term in the sum.  Thus,

$$\frac{1}{n^2}\sum_{j=2}^n \bar{d}(j)\sim O\left(\frac{1}{n}\right).$$

Finally then, using convexity on $(\delta+\sqrt{n\log{n}})^2\leq \delta^2+n\log{n}$,

$$
\begin{aligned}
\label{empiricalcomp}
\frac{d\left(\widehat{W}_s,W_{true}\right)^2}{d\left(\widetilde{W}_{obs},W_{true}\right)^2} \sim O\left( \left(\frac{\delta}{n}\right)^2 + \frac{\log(n)}{n}  \right).
\end{aligned}
$$

If $$\delta=\bar{d}(1)\sim o\left(\sqrt{n\log(n)}\right)$$ this becomes $O\left( \frac{\log(n)}{n}\right).$

\begin{flushright}
$\square$
\end{flushright}

\section{Proof of Theorem 4}
\label{C}

\begin{enumerate}

\item  The below results concerning the lack of continuity of various statistics in any matrix norm are based on the following simple observation.  If the statistic uses the geodesic distance between two vertices in its definition, it cannot be continuous in any matrix norm, as we show more precisely below. 

\begin{enumerate}
\item  For any two vertices $x,y$, $$d_G(x,y)=\inf \{ k : W^k(x,y)>0\}.$$  The following example illustrates how to proceed in the general case.  

Consider a dumbbell graph with adjacency matrix $W$ with a single edge, $e$ connecting the two bells.  Define $W_{\epsilon}$ to be this graph with with $W(e)=\epsilon$.  Since all norms on $\mathbb{R}^n$ are equivalent, it's clear that in any norm, $W_{\epsilon}\rightarrow W_0$ as $\epsilon\downarrow 0$ but that $d_{W\epsilon}(x,y)$ is constant for any $\epsilon>0$ for any $x$ in one bell and $y$ on the other and $d_{W_0}(x,y)=\infty$ so that $d_{G}$ isn't continuous in $G$.

In the general case, restrict to a connected graph, and two vertices $x$ and $y$.  If there are many geodesics of the same length connecting $x$ and $y$, proceed as above perturbing by $\epsilon$ as many edges necessary to sever these geodesics.  The argument is the same.

\item  $$C_B(v)=\sum_{s\neq v\neq t\in V}\frac{\sigma_{s,t}(v)}{\sigma_{s,t}}$$ where $\sigma_{s,t}$ are the number of geodesics connecting $s$ and $t$ and $\sigma_{s,t}(v)$ are the number of these geodesics passing through $v$.  From (1) of this theorem, geodesic distances aren't continuous functions of $W$, so neither is $\sigma_{s,t}$, and hence, $\sigma_{s,t}(v)$.

\item  $$C_C(v)=\frac{\sum_{t\in V-v} d_G(v,t)}{n-1}$$ where $|V|=n$.  Again from (1) of this theorem, geodesic distances aren't a continuous function of $W$, so we can apply the methods of (1) directly to this quantity to conclude that $C_C$ isn't continuous in $W$ in any norm.  
\end{enumerate}

\item  The ''in any norm" part is a consequence of the fact that any two matrix norms are equivalent \citep{r1}.  Thus, we may show the results for any particular matrix norm.

\begin{enumerate}

\item  $$C_D(W,x)=deg(x)$$ where $deg(x)=\sum_{y} W(x,y)$.  So,

\begin{multline}
\left|C_D(W,x)-C_D(W',x)\right| = \left| \sum_y W(x,y)-\sum_y W'(x,y)\right|\\ \nonumber
 \leq \sum_y \left|W(x,y)-W'(x,y) \right| \\
 \leq \max_x \left( \sum_y \left| W(x,y)-W'(x,y)\right| \right):= |||W-W'|||_1\nonumber
 \end{multline} 

\item  Let $W_1=W_0+E$ with $||W_1-W_0||=|m|<\epsilon$.  Using the binomial theorem, and some algebra,

\begin{multline}
||W_1^k-W_0^k|| \leq \sum_{\ell=1}^k {k\choose \ell} |m|^{\ell}||W_0||^{k-\ell}\\
=|m|\sum_{\ell=0}^{k-1} {k-1\choose \ell} \cdot \frac{k}{\ell+1} |m|^{\ell} ||W_0||^{k-1-\ell}\\ \nonumber
\leq k|m| \left( |m|+||W_0||\right)^{k-1}. \nonumber
\end{multline}

Thus,

$$||W_1^k-W_0^k||\leq k|m| ||W_0||^{k-1} \left( 1+\frac{|m|}{||W_0||}\right)^{k-1}.$$

\item  To simplify notation, fix $S\subset V$ and define $$f(W)=\sum_{x\in S,y\in S^c}W(x,y)$$ and $$g(W)=\min\{Vol S,VolS^c\}.$$

First, note that $f$ and $g$ are Lipschitz.

$$
\begin{aligned}
 |f(W_1)-f(W_2)|&=\left|\sum_{x\in S,y\in S^c}W_1(x,y) - \sum_{x\in S,y\in S^c}W_2(x,y)\right| \\
& \leq \sum_{x\in S,y\in S^c}\left| W_1(x,y)-W_2(x,y)\right|\\
& \leq ||W_1-W_2||_1
\end{aligned}
$$

For $g$, since

$$g(W)=\frac{1}{2} \left[VolS+VolS^c-\left| VolS - VolS^c\right| \right]$$

we have

\begin{multline}
g(W_1)-g(W_2) =\frac{1}{2} (Vol_1S+Vol_1S^c-\left| Vol_1S - Vol_1S^c\right|\\
-\left[Vol_2S+Vol_2S^c-\left| Vol_2S - Vol_2S^c\right| \right] )
\nonumber
\end{multline}

so that grouping terms and using the reverse triangle inequality,

$$|g(W_1)-g(W_2)|\leq 2||W_1-W_2||_1.$$

So, when $S$ contains at least one edge of weight $\delta>0$, $$\min\{g(W_1),g(W_2)\}>\delta,$$

$$
\begin{aligned}
&\left|\phi_S(W_1)-\phi_S(W_2)\right| \\
&=\left|\frac{f(W_1)}{g(W_1)}-\frac{f(W_2)}{g(W_2)}\right|\\
&=\frac{1}{g(W_1)g(W_2)}\left| g(W_2)f(W_1)-f(W_2)g(W_1)\right|\\
&\leq \frac{1}{g(W_1)g(W_2)} g(W_2)\left[\left|f(W_1)-f(W_2)\right| \right.\\
&\hspace{1.5in} \left.+ f(W_2)\left|g(W_1)-g(W_2)\right|\right]\\
&\leq \frac{||W_1-W_2||_1}{g(W_1)}\left[ 1+ 2\frac{f(W_2)}{g(W_2)}\right]\\
&\leq \frac{3}{\delta} ||W_1-W_2||_1
\end{aligned}
$$

\item

If $W$ is the adjacency matrix of a connected, undirected, network then it is symmetric.  By the spectral theorem, the eigenvalues of $W$ are real valued, and there exist an orthonormal basis of eigenvectors.  Order the eigenvalues in descending order, and let $v$ be the unique, positive, eigenvector with largest eigenvalue with $L^2$ norm 1 (by the Perron-Frobenius theorem).  The value $v(i)$ at vertex $i$ is the eigenvector centrality of vertex $i$.

Now, let $W_1$ and $W_2$ be two such matrices, with eigenvector centrality $v_1$ and $v_2$.  Define,

$$d(v_1,v_2)=\sqrt{1-\left| \left<v_1,v_2\right> \right|^2}$$ which is the sine of the angle between $v_1$ and $v_2$.

\begin{prop}
\label{eigcentmet}
$d(v_1,v_2)$ is a metric on the set of all eigenvector centraliities.
\end{prop}

{\bf Proof of Proposition \ref{eigcentmet}:}  

Symmetry is obvious, and is positiveness.  Definiteness comes from the fact that if $d(v_1,v_2)=0$, then $v_1$ and $v_2$ are parallel, but being completely positive and having $L^2$ norm 1, they must be equal.

The triangle inequality is a little more subtle:

$$
\begin{aligned}
d(a+b,c)^2&=1-\left|\left<a+b,c\right>\right|^2\\
&=1-\left|\left<a,c\right>\right|^2-\left|\left<b,c\right>\right|^2-2\left<a,c\right>\left<a,b\right>\\
&=d(a,c)^2+d(b,c)^2-2\left<a,c\right>\left<a,b\right>-1\\
&=(d(a,c)+d(b,c))^2-2\left<a,c\right>\left<a,b\right>\\
&\hspace{1.5in}-1-2d(a,c)d(b,c)\\
&\leq (d(a,c)+d(b,c))^2\\
\end{aligned}
$$

where we've used that $<a,c>,<a,b>\geq 0$. 

\begin{flushright}
$\square$
\end{flushright}

We now need an important theorem regarding eigenvector perturbation known as the Davis-Kahan Theorem \citep{r2},\citep{r3}.

\begin{prop} (Davis-Kahan)  Let $A$ and $B$ be two symmetric matrices.  For any such matrices, let $\lambda(A)$ and $\lambda(B)$ denote their eigenvalues.  For any subset $S\subseteq \mathbb{R}$, let $$\delta=\min_{\lambda \in \lambda(A)\cap S^c, s\in S}\{ | \lambda-s|\}.$$  Let $X_A$ be an orthonormal matrix whose column space is equal to the eigenspace of $A$ corresponding to eigenvalues in $\lambda_S(A)$ and similarly for $X_B$.  Then,

$$d(X_A,X_B)\leq \frac{||A-B||_F}{\delta}$$ where $||\cdot ||_F$ denotes the Frobenius norm, and $d(X_A,X_B)$ is the Frobenius norm of the singular values of $X_A'X_B.$
\end{prop}

To prove the result then, take $A$ and $B$ be two adjacency matrices for two undirected, connected, networks.  Then,

$$d(v_A,v_B)\leq \frac{||A-B||_F}{\delta}.$$

\item  If there are $E(W)=\frac{1}{2}\sum_{i,j}W(i,j)$ edges in a network $W$, the density is defined as,

$$\mathfrak{d}(W)=\frac{E(W)}{{n\choose 2}}.$$

Thus,

$$|\mathfrak{d}(W_1)-\mathfrak{d}(W_2)|\leq \frac{ ||W_1-W_2||_1}{{n\choose 2}}.$$

\end{enumerate}
\end{enumerate}

\begin{flushright}
$\square$
\end{flushright}

\end{enumerate}	

\end{document}